\documentclass[manuscript]{acmart}

\usepackage{algorithm}
\usepackage{algorithmic}
\usepackage{graphicx}
\usepackage{subcaption}
\usepackage{amsmath}
\usepackage{cancel}
\usepackage[normalem]{ulem}
\AtBeginDocument{%
  }

\setcopyright{acmcopyright}
\copyrightyear{2018}
\acmYear{2018}
\acmDOI{XXXXXXX.XXXXXXX}

\acmConference[Conference acronym 'XX]{Make sure to enter the correct
  conference title from your rights confirmation emai}{June 03--05,
  2018}{Woodstock, NY}
\acmPrice{15.00}
\acmISBN{978-1-4503-XXXX-X/18/06}

\begin{document}

\title{Subgoal-Based Explanations for Unreliable Intelligent Decision Support Systems}

    
\author{Devleena Das}
\email{ddas41@gatech.edu}
\affiliation{%
  \institution{Georgia Institute of Technology}
   \country{USA}
}
\author{Been Kim}
\email{beenkim@google.com}
\affiliation{%
  \institution{Google Research}
   \country{USA}
}
\author{Sonia Chernova}
\email{chernova@gatech.edu}
\affiliation{%
  \institution{Georgia Institute of Technology}
   \country{USA}
}

\renewcommand{\shortauthors}{Das et al.}

\begin{abstract}
Intelligent decision support (IDS) systems leverage artificial intelligence techniques to generate recommendations that guide human users through the decision making phases of a task. However, a key challenge is that IDS systems are not perfect, and in complex real-world scenarios may produce suboptimal output or fail to work altogether. The field of explainable AI (XAI) has sought to develop techniques that improve the interpretability of black-box systems. While most XAI work has focused on single-classification tasks, the subfield of explainable AI planning (XAIP) has sought to develop techniques that make sequential decision making AI systems explainable to domain experts. Critically, prior work in applying XAIP techniques to IDS systems has assumed that the plan being proposed by the planner is always optimal, and therefore the action or plan being recommended as decision support to the user is always optimal. In this work, we examine novice user interactions with a non-robust IDS system -- one that occasionally recommends suboptimal actions, and one that may become unavailable after users have become accustomed to its guidance.  We introduce a new explanation type, \textit{subgoal-based explanations}, for plan-based IDS systems, that supplements traditional IDS output with information about the subgoal toward which the recommended action would contribute. We demonstrate that subgoal-based explanations lead to improved user task performance in the presence of IDS recommendations, improve user ability to distinguish optimal and suboptimal IDS recommendations, and are preferred by users. Additionally, we demonstrate that subgoal-based explanations enable more robust user performance in the case of IDS failure, showing the significant benefit of training users for an underlying task with subgoal-based explanations.

% are preferred by users, and enable more robust user performance in the case of IDS failure. 
\end{abstract}

\begin{CCSXML}
<ccs2012>
   <concept>
       <concept_id>10010147.10010178.10010199</concept_id>
       <concept_desc>Computing methodologies~Planning and scheduling</concept_desc>
       <concept_significance>500</concept_significance>
       </concept>
 </ccs2012>
\end{CCSXML}
\begin{CCSXML}
<ccs2012>
 <concept>
<concept_id>10003120.10003121.10003124</concept_id>
<concept_desc>Human-centered computing~Interaction paradigms</concept_desc>
<concept_significance>500</concept_significance>
</concept>
</ccs2012>
\end{CCSXML}

\ccsdesc[500]{Human-centered computing~Interaction paradigms}
\ccsdesc[500]{Computing methodologies~Planning and scheduling}

\keywords{Explainable AI, Intelligent Decision Support Systems, Planning}

\maketitle

\section{Introduction}
Intelligent decision support (IDS) systems leverage artificial intelligence techniques to generate recommendations that guide human users through the decision making phases of a task \cite{sutton2020overview}. While much prior work has focused on decision support for domain experts (e.g., cancer diagnosis for oncologists \cite{walsh2019decision}), increasingly, IDS systems have been proven particularly useful in helping \textit{novice users} make decisions \cite{Gutierrez2019,machado2018use}. However, a key challenge is that IDS systems are not perfect, and in complex real-world scenarios the actions recommended by IDS systems may be far from optimal \cite{guerlain2000intelligent}.  Such errors particularly affect novice users, who lack the knowledge to assess the correctness of an IDS recommendation \cite{nourani2020role}.

The field of explainable AI (XAI) has sought to make the decision making of AI systems more transparent by developing interpretability techniques for black-box AI models \cite{doshi2017towards}. While prior work on XAI largely focuses on explaining single classification tasks, the subfield of \textit{explainable AI Planning (XAIP)} seeks to explain decisions in sequential decision making tasks. Prior work in XAIP has largely focused on explaining plan solutions that help users answer questions ``Why plan P?" and "Why not plan Q?" through techniques such as causal-link-chain (CLC) explanations \cite{seegebarth2012making} and contrastive explanations \cite{hoffmann2019explainable}. These techniques have been effective in helping domain-experts understand how their proposed solution differs from a planner's solution \cite{hoffmann2019explainable}, as well as how performing a current action affects the preconditions of future actions \cite{seegebarth2012making}.
% The field of \textit{explainable AI Planning (XAIP)} have developed techniques to make sequential decision making systems more understandable to domain-experts.

% Critically, prior work in applying XAIP techniques to IDS systems has assumed that the plan being proposed by the planner is always optimal, and therefore the action or plan being recommended to the user is always optimal \cite{grover2020radar, valmeekam2020radar}. However, optimal IDS decision making cannot be guaranteed in complex real world deployments. In fact, in real world systems, other assumed, robust characteristics of IDS systems may not hold true, including the ability to always receive suggestions at deployment.
Critically, prior work in applying XAIP techniques to IDS systems has assumed that the planner within the plan-based IDS system is always optimal, and therefore the actions being recommended to the user are always optimal \cite{grover2020radar, valmeekam2020radar}. However, the assumption of robust IDS system, with an optimal planner, cannot be guaranteed in complex real world deployments. In fact, in real world systems, other assumed, robust characteristics of IDS systems may not hold true as well, such as the ability to always receive suggestions at deployment. In other words, there may be situations in which a user's query is unanswerable, or the IDS system runs into a failure and is no longer available to the user. 

In this work, we examine novice user interactions with a non-robust IDS system -- one that occasionally recommends a suboptimal action, and one that may become unavailable after users have become accustomed to its guidance.  A user of such a system, given an IDS action recommendation, must be able to determine whether the recommendation is optimal or not.  Additionally, in the absence of an IDS recommendation, the user should have sufficient understanding of the task such that their task performance is not negatively impacted by the sudden absence of previously available recommendations.
Leveraging insights from psychology, which demonstrate that humans naturally break down large complex tasks into a smaller set of more manageable subgoals \cite{newell1972human, csibra2007obsessed, vallacher1987people, schank2013scripts}, we introduce a new explanation type -- \textit{subgoal-based explanations} -- that supplements traditional IDS output with information about the subgoal toward which the recommended action would contribute. We then evaluate the impact such an explanation has on novice user performance through experiments with 105 participants in a complex restaurant planning domain. We compare our subgoal-based explanations with the traditional action recommendation outputs of IDS systems, as well as causal-link chain (CLC) explanations \cite{seegebarth2012making}, an XAIP technique most relevant to our work. We contribute several key findings:

\begin{itemize}
    \item In the context of a suboptimal IDS system, subgoal-based explanations enable users to successfully detect and avoid more suboptimal IDS recommendations than users who are only provided traditional IDS action recommendations or CLC explanations.
    
    \item Users who receive subgoal-based explanations achieve better performance when performing a task under IDS guidance than users who receive IDS guidance with CLC or those who receive traditional IDS action recommendations.

    \item Users who are exposed to subgoal-based explanations for some period of time, are able to perform the task more reliably in the absence of further IDS, compared to users who only receive action recommendations or CLC explanations.  This finding suggests that explanations contribute a significant training benefit beyond both traditional IDS output as well as CLC explanations.
     
    \item In a direct comparison, users exhibit a strong preference for IDS output that includes subgoal-based explanations versus CLC explanations or IDS output that traditionally only includes the next action.
\end{itemize}
We also show a simple way to generate domain-independent subgoal-based explanations that can generalize to any hierarchical plan-based system and are broadly applicable across a wide range of IDS systems and application areas. Together with our findings, our work is a first step towards investigating how and when subgoal-based explanations in XAIP systems are beneficial in complex real-world IDS systems that are not fail-proof.

% First, we introduce a new explanation type within IDS systems that leverage planning, known as subgoal-based explanations. Second we validate that while under optimal conditions both SB and regular action-based support from IDS systems lead to near-optimal user performance, that in the presence of noisy advice SB explanations enable the user to more effectively distinguish bad advice from good advice. Third, we demonstrate that SB explanations more effectively improve user understanding of the underlying task, such that once IDS is removed entirely, users who received prior SB explanations are able to significantly outperform those who received action-based support. Finally, we show that SB explanations do not only provide quantitative improvements in task accomplishment, but also are qualitatively more preferred by end users than action-based support.  

\section{Related Work}
In this section, we situate our work in the context of the two prominent research areas most closely related to our work: Intelligent Decision Support Systems and Explainable AI.

\subsection{Intelligent Decision Support (IDS) Systems}
IDS systems have been developed to assist domain-experts as well as novice-users in decision making across a wide range of applications. For example, in the context of aiding domain-experts, IDS systems have been investigated in clinical settings such as aiding doctors in prescribing medicine \cite{lin2009decision} or aiding general practitioners for pathology orderings \cite{zhuang2009combining}. Similarly, in the context of aiding novice-users, IDS systems have been developed to help dental students manage dental trauma \cite{machado2018use}, aid non-financial experts in stock market investment \cite{Gutierrez2019}, and support novice business owners in management decisions \cite{arnold2004impact}.
Among these decision support systems, depending on the domain, there are a range of mediums through which support is provided. For instance, in Machado et al. \cite{machado2018use}, the authors develop a mobile app for clinical decision support that allows dental students to answer a series of questions to determine a diagnosis and provide treatment suggestions. Similarly, in Gutierrez et al. \cite{Gutierrez2019}, researchers investigate how to best portray visual representations and interaction techniques to aid novice users in business decisions. Alternatively, Papamichail \& French \cite{papamichail2000decision} develop a natural language generator to justify decision support in nuclear emergencies via natural language based reports.

Given that many IDS systems interact with end-users who are not AI-experts, several bodies of work have investigated how to enhance the transparency of IDS systems to improve user trust. These transparency techniques have been primarily studied in the context of explaining recommendations for single-classification tasks, such as
clinical decision support \cite{jones2019malfunction, feng2020explainable}. For example, Jones et al. \cite{jones2019malfunction} develop a classification model that can be used in the testing process of a clinical decision support system to identify failure modes. Additionally, Feng et al. \cite{feng2020explainable} leverage ClinicalBERT and attention mechanisms to provide interpretable rationales for predicting Sepsis and mortality in ICUs. By contrast, our work investigates IDS in sequential decision making settings. Specifically, we investigate how to provide explanations that help novice users improve their decision making performance in the presence of potentially suboptimal suggestions.

% IDS systems have been developed to assist users in decision making across a wide range of applications, such as providing assistance to domain-experts in clinical settings \cite{walsh2019decision, zhuang2009combining} and aiding novice-users in management settings \cite{machado2018use, Gutierrez2019}. Amongst these IDS systems, decision support is provided through a range of mediums, depending on the domain. 
% For instance, in \cite{machado2018use}, the authors develop a mobile app for clinical decision support that allows dental students to answer a series of questions to determine a diagnosis and provide treatment suggestions. In \cite{Gutierrez2019}, researchers investigate how to best portray visual representations and interaction techniques to aid novice users in business decisions. 

% Given that many IDS systems interact with end-users who are not AI-experts, several bodies of work have investigated how to enhance the transparency of IDS systems to improve user trust. These transparency techniques have been studied in the context of explaining recommendations for single-classification tasks, such as
% clinical decision support to identify failure modes \cite{jones2019malfunction, feng2020explainable}. By contrast, our work investigates IDS in sequential decision making settings. Specifically, we investigate how to provide explanations that help novice users improve their decision making performance in the presence of potentially suboptimal suggestions.

\subsection{Explainable AI}
The field of explainable AI (XAI) aims to improve a user's understanding of the inner workings of complex models \cite{doshi2017towards, adadi2018peeking}. Given that AI and ML models are not guaranteed to be optimal, an important objective of XAI techniques includes being able to help users identify vulnerabilities or ``bugs" within a model \cite{adebayo2020debugging} as well as identify any spurious correlations \cite{ kim2018interpretability}. Our work explores a scenario in which the underlying AI model, and therefore the explanation that results from it, may be suboptimal.  We examine how, even under these settings, users can leverage explanations to ultimately improve their task performance. 

While the field of XAI has primarily focused its interests in interpretability techniques for classification-based tasks to improve AI-expert understanding \cite{zhang2019interpreting, ribeiro2016should, kim2018interpretability}, a recently growing field of XAI research has focused on explainability techniques for sequential-decision making systems \cite{chakraborti2020emerging}. Specifically, the subfield of explainable AI Planning (XAIP) seeks to develop methods for explaining sequential decision making problems, where an agent engages in a longer-term interaction with a user \cite{chakraborti2020emerging}. To provide greater context for our work, we first review the types of explanations that have been developed in sequential decision making. Within the community, techniques have primarily focused on explaining an agent's entire plan solution to end-users. A recent survey by Chakraborti et al. \citep{chakraborti2020emerging} highlights the key areas of plan explanations that have been investigated, including generating contrastive explanations~\cite{hoffmann2019explainable}, explaining unsolvable plans~\cite{sreedharan2019can}, and generating explicable plan explanations \cite{chakraborti2017plan}. 

Additionally, XAIP techniques have also been applied to plan-based decision support systems in efforts to improve human-in-the-loop planning. For example, RADAR by Grover et al. \cite{grover2020radar} provides XAIP features such as plan summarization, plan explanations in the form of minimally complete contrastive explanations, plan validation, and action and plan suggestions to improve domain-expert decision making. Furthermore, Valmeekam et al.\cite{valmeekam2020radar} develop RADAR-X which leverages user queries to understand user preferences for providing refined plan suggestions. Our work similarly aims to support human-in-the-loop planning, with important differences in that we do not make the assumption of an optimal AI planner or optimal IDS system recommendations. Specifically, we intentionally examine the role of plan-based IDS systems in the context of suboptimal suggestions, and our work focuses on providing explanations of IDS systems to novice users.

Furthermore, under sequential decision making scenarios, growing bodies of work have investigated how to provide meaningful explanations to end-users and what types of information aid non-AI expert understanding. For example, Ehsan et al. \cite{ehsan2019automated} utilize sequence to sequence learning to autonomously translate state and action information of a game agent into natural language. In a user study, the authors validate a wide range of human factors that influence users' preferences of explanations, concluding that users prefer "complete-view" rationales which utilized the entire game state space as context. Similarly, Das et al. \cite{das2021explainable} utilize machine translated, natural language explanations to help non-AI experts understand robot failures and provide accurate recovery. Results from Das et al. \cite{das2021explainable} demonstrate that contextual information from the environment significantly improves users' failure understanding in comparison no contextual information. Additionally, within the planning domain, Canal et al. \cite{canal2021task} propose PlanVerb, a domain and planner independent method for verbalizing task plan solutions into natural language to increase user understandability of plan solutions. Using PlanVerb, the authors generate natural language plan explanations by semantically tagging action and predicate information. In our work, we seek to provide investigate how contextual information in the form of \textit{subgoals} in IDS systems can help novice users improve their decision making in the presence of non-robust IDS systems.

\section {Problem Formulation}
In this section, we first provide definitions for the planning problem that underlies our IDS system.  We then formalize the problem of providing explanations in support of plan-based IDS and present our research hypotheses.
%within the context of a plan-based IDS system in which the underlying decision-making model is a planner. We first provide background on how a planning problem is solved. We then formulate how explanations supplementing a plan-based IDS system's recommendations can benefit a novice user's task performance.

%plan-based IDS systems
%Specifically, we first define the underlying planning problem solved within a plan-based IDS system. We then describe how explanations from plan-based IDS systems can help users reduce automation bias. 

%the decision making pipeline of a plan-based IDS system. We first define the underlying planning problem solved within the IDS system, and then we formulate how the plan solution is utilized by the IDS system to provide recommendations.

\subsection{Planning Problem}
\label{sec:planning-problem}

Formally, a classical planning problem \cite{norvig2002modern} is defined by a model $\mathcal{M}$ = $\langle\mathcal{D},\mathcal{I},\mathcal{G}\rangle$ where domain $\mathcal{D}$ is represented by $\langle F,A \rangle$, such that $F$ is a finite set of fluents that define a state $s\subseteq F$, and $A$ represents a finite set of actions. $\mathcal{I}$ and $\mathcal{G}$ represent the initial and goal states, respectively, such that $\mathcal{I}$,$\mathcal{G}$ $\subseteq F$.  Note that $\mathcal{G}$ may be modeled as a set of $\langle g_{0}...g_{j}\rangle$, where $g_i \in \mathcal{G}$ represents a subgoal.  An action $a \in A$ is defined by a tuple $\langle c_{a}, pre(a), eff^{+}(a), eff^{-}(a)\rangle$, where $c_{a}$ is the associated cost of $a$, and $pre(a), eff^{+}(a), eff^{-}(a)$ denote the set of preconditions, add and delete effects, respectively. An action $a \in A$ can only be executed in a state $s$ if $s \models pre(a)$. A transition function, $\delta_{\mathcal{M}}(s,a)$ is used to transition an agent from $\mathcal{I}$ to $\mathcal{G}$, performing a sequence of actions $\langle a_{1}...a_{n}\rangle$, each with an associated cost $c_{a}$. In other words, the cost of plan $C(\pi, \mathcal{M})$ is defined by $\sum_{a \in \pi} c_{a}$, the sum cost of all actions within the plan, or $\infty$ if the goal is not met. The solution to a planning problem is a plan $\pi = \langle a_{1}...a_{n}\rangle$ such that $\delta_{M}(\mathcal{I},\pi \models \mathcal{G}$). The optimal plan solution, $\pi^*$, is defined by $argmin_{\pi} \{C(\pi,\mathcal{M}) \forall \pi$ such that $\delta_{\mathcal{M}}(\mathcal{I}, \pi) \models \mathcal{G}\}$. 

In general, the objective of a classical planning problem is to solve for a plan solution, made up of a sequence of actions that transforms the AI agent from its initial state to a goal state. Each action within a plan solution has an associated cost, and may include a set of preconditions that need to be met. Each action, when performed, also produces a set of effects within the domain. An optimized plan solution is one that has the lowest plan cost. Below we formalize the problem of providing explanations for action suggestions from plan-based IDS systems.

\subsection{Explainability in Plan-Based IDS Systems}
\label{sec:explainability-plan-based-ids}
The goal of a plan-based IDS system is to provide the user with action recommendations $a_{IDS} \in \pi$.  In turn, the user, who is given $a_{IDS}$ as input, must select their own action $a_h$ to take in response. In the ideal case, the IDS guides the user along some optimal plan $\pi^*$ by always recommending an optimal action, $a_{IDS} = a^* \in \pi^*$, which results in the user always taking the optimal action, $a_{h} = a^* \in \pi^*$.  However, there are two limitations to this idealized formulation.

First, in complex real-world scenarios an IDS systems may not be able to generate an optimal plan, resulting in suboptimal recommendations \cite{guerlain2000intelligent}.  In this case, the user is faced with the challenge to discern whether the IDS system's recommended action is optimal (i.e., $a_{IDS} \overset{?}{=} a^*$).  Relating to this, we state the following hypothesis:
\begin{quote}
\noindent \textbf{H1:} We hypothesize that there exists a type of explanation, $\mathcal{E}$, that when presented in conjunction with $a_{IDS}$ can aid users in determining $a_{IDS} \stackrel{?}{=} a^*$. 
\end{quote}
Specifically, we hypothesize that with the aid of $\mathcal{E}$, users will more likely accept 
$a_{IDS}$ when $a_{IDS} = a^*$ and correctly reject $a_{IDS}$ when $a_{IDS} \neq a^*$. 

Second, in complex real-world scenarios an IDS system may not always be available due to being offline, a failure, or the query being outside its scope.  In this case, the user may suddenly be required to select $a_h$ without the benefit of an IDS system's guidance. This scenario will pose a particularly significant challenge to users who had previously only performed the task under the support of an IDS system.
Relating to this, we state the following hypothesis:
\begin{quote}
%\textbf{H2:} We hypothesize that exposure to explanations $\mathcal{E}$ serves as a form of \textit{training} for users,
%will provide the user with deeper understanding of the task, 
%such that future user task performance in the absence of IDS assistance will be greater than the performance of users who were not exposed to $\mathcal{E}$.
\textbf{H2:} We hypothesize that exposure to explanations $\mathcal{E}$ improves user understanding of the task, such that when IDS recommendations are turned off, users with previous exposure to $\mathcal{E}$ will achieve greater task performance than users who had the same amount of domain experience but without exposure to $\mathcal{E}$.  
\end{quote}
Specifically, we posit that users previously exposed to explanation $\mathcal{E}$ will be able to select actions $a_h$ that lead to more optimal task performance than users who were not previously exposed to explanation $\mathcal{E}$ (i.e, $C(\pi_{h}^{\mathcal{E}}, \mathcal{M}) < C(\pi_{h}^{\cancel{\mathcal{E}}}, \mathcal{M})$).  In this perspective, explanations can be seen as a \textit{training mechanism} that leverages IDS to improve user understanding of the task.
%given only traditional IDS output.

Finally, prior work across many XAI applications has demonstrated that incorporating explanations into the output of automated systems improves user performance in a given task \cite{das2020leveraging,tabrez2019explanation}.  
Relating to this, we state the following two hypotheses in the context of IDS systems:
\begin{quote}
\textbf{H3:} We hypothesize that user performance on the task will improve when IDS output, $a_{IDS}$, is supplemented with explanation $\mathcal{E}$.  \\
\textbf{H4:} We hypothesize that users will prefer the output of a system that includes $\mathcal{E}$ over a system that only includes $a_{IDS}$.
\end{quote}
Specifically, we posit that overall plan cost for users exposed to explanations will be lower than for users who did not receive explanations (i.e., $C(\pi_{h}^{\mathcal{E}}, \mathcal{M}) < C(\pi_{h}^{\cancel{\mathcal{E}}}, \mathcal{M})$), and that users will prefer to see explanations as part of an IDS output.

In the next sections, we first present a new variant of $\mathcal{E}$ for explaining IDS systems action recommendations, leveraging findings in psychology \cite{newell1972human, csibra2007obsessed, vallacher1987people, schank2013scripts}.  We then describe our validation domain, and the experiments that were conducted to support the above hypotheses.

\section{Subgoal-Based Explanations}
Research in psychology shows that humans faced with a complex sequential decision making task naturally construct a mental model of the task as a decomposition of multiple subgoals \cite{newell1972human, csibra2007obsessed, vallacher1987people, schank2013scripts}. The utility of such subgoals has been briefly explored within the XAIP and AI community. For example, many AI techniques utilize hierarchical structures to leverage the improved computational efficiency of such representations \cite{iovino2020survey,kaelbling2010hierarchical}. Similarly, subgoals have been leveraged to explain the infesability of a defined planning problem \cite{sreedharan2019can}, and also learned via human annotations to label optimal plan solutions to humans in an interpretable manner \cite{zhang2017plan}. These approaches, however, have explored the utility of subgoals in extreme scenarios, to explain unsolvable planning definitions or to explain plan solutions that are always optimal. As mentioned in Section \ref{sec:explainability-plan-based-ids}, in more complex, real-world scenarios, the reliability of IDS systems may be more \textit{subtle}. In other words, a plan-based decision support may \textit{sometimes} provide suboptimal solutions, and \textit{sometimes} become unavailable to end-users who may be depending on the IDS systems. Therefore, in this work, inspired by the natural hierarchical representations used by both human users and AI systems, we introduce a new type of explanation known as subgoal-based explanations, $\mathcal{E_{SB}}$ which aims to improve user task performance both in optimal and suboptimal IDS settings.
Below we further detail the definition of $\mathcal{E_{SB}}$ by leveraging the definition of a planning problem (see Section \ref{sec:planning-problem}): 
\begin{quote}
    \textbf{$\mathcal{E}_{SB}$}: Given planning goal $\mathcal{G}$, which is decomposed into a set of subgoals $\langle g_{0}...g_{n}\rangle$, a subgoal-based explanation is described by $\mathcal{E}_{SB} =  a_{IDS} + g_{i}(a_{IDS})$. 
\end{quote}
In other words, a subgoal-based explanation provides the next recommended action from an IDS system, $a_{IDS}$ along with the associated subgoal $g_{i}$ that is satisfied by $a_{IDS}$. For example, consider a car manufacturing scenario where the goal, $\mathcal{G}$, is building a car. The overall goal requires multiple subassemblies which can be formulated as subgoals, $\langle g_{0}...g_{n}\rangle$. Subassemblies may include chassis assembly, drivetrain assembly, car interior assembly and external assembly for the car. In this scenario, examples of subgoal explanations, $\mathcal{E}_{SB}$, may include ``Attach the lights to finish the internal assembly," where ``internal assembly" is the subgoal. Similarly another explanation may include, ``Fasten the bolts for the chassis before fastening bolts for the drivetrain", where ``for the chassis" and ``for the drivetrain" are the subgoals. We posit that especially in scenarios of lengthy task plans, with often overlapping actions (i.e fasten the bolts) for different requirements (subgoals), $\mathcal{E}_{SB}$ explanations can help end-users verify the IDS system's underlying reasoning for a particular action. Understanding such reasoning is crucially important in the context where IDS systems may \textit{sometimes} be suboptimal. In general, subgoals for a particular plan may be predefined in the planning representation, or may be autonomously derived \cite{richter2008landmarks,czechowski2021subgoal, zhang2017plan}.  In this work, we encode the subgoals within our planning problem definition.

%An example of a $\mathcal{E}_{SB}$ is ``Chop the tomato to prepare the lasagna," where ``to prepare the lasagna" is the subgoal.

\section{IDS in Restaurant Planning Domain}
\label{Restuarant IDS}
We investigate the efficacy of $\mathcal{E_{SB}}$ explanations in the context of a complex sequential planning scenario: running a restaurant kitchen. In our task, a user plays as a chef to deliver a set of $M$ meals, each within unique delivery times, with the help of an anthropomorphized IDS known as Manager Molly. Below, we further detail the restaurant planning game and our plan-based IDS system.

\subsection{Restaurant Game Overview}

\begin{figure*}[b]
\centering
\vspace{-.5cm}
\includegraphics[width=10cm]{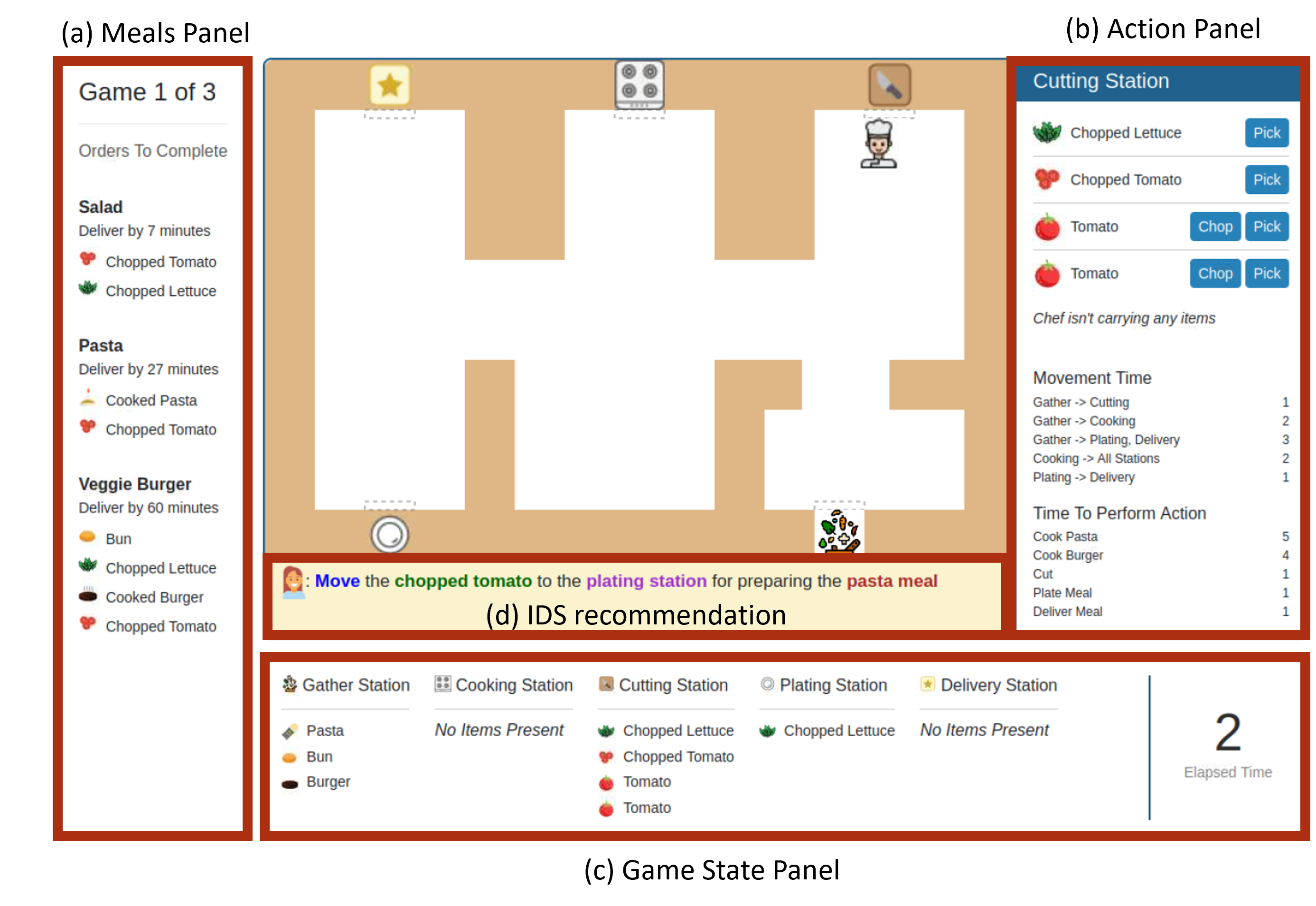}
\caption{A visualization of our restaurant planning game interface that highlights, via the red boxes, the different types of information available to the user at any given time, including (a) the set of meals that need to be delivered and their respective delivery times, (b) the set of actions available at the user's current station, (c) the game state information, and (d), the provided IDS output.}
\label{fig:game_screen}
\end{figure*}

Figure \ref{fig:game_screen} provides a visualization of our restaurant planning game which is inspired by the online game Overcooked. Variations of the Overcooked restaurant planning game have been widely studied in the AI community because tasks in this domain are difficult for users to optimize \cite{carroll2019utility,wu2021too,liu2020planning}. Particularly, characteristics such as parallelizable actions (i.e cooking the salmon, while chopping the tomato), variable action durations, and one-to-many mapping between actions and the subgoal they achieve (i.e multiple dishes may require chopping tomatoes) all provide challenges for both users and IDS systems. Therefore, we posit that identifying suboptimal suggestions in such domain is non-trivial since a single action can satisfy multiple subgoals (lettuce for salad vs. burger) and parallelizing tasks can lead to surprising orderings (boiling water for pasta while preparing salad). These characteristics especially motivate us to investigate the utility of subgoal based explanations under a similarly designed restaurant planning game.

Within our game, the user controls a chef avatar, and utilizes five meal prep stations to prepare $M$ meals consisting of various ingredients. The game objective is to deliver the meals to restaurant customers within the designated meal delivery time for each meal, denoted as $t^m_{goal\_delivery}$. As seen in Figure \ref{fig:game_screen},  panel (a) in the game interface illustrates the set of meals the user has to complete in a given game, along with their associated delivery times. The five meal-prep stations include: \textit{gather station}, \textit{cutting station}, \textit{cooking station}, \textit{plating station}, and \textit{delivery station}. Within the game, the user is able to perform an action $a_{h}$ from the action space $A = \{\textit{cut, move-chef, move-item, start-cook, end-cook, deliver}$, $\textit{prepare-meal}\}$, so long as the preconditions of $a_{h}$ are met and that the effects of $a_{h}$ result in a valid game state $S$. The game does not allow players to perform an invalid action (e.g., start cooking the steak without the steak being present at the cooking station). Panel (b) in the game interface provides the subset of actions, $A_{station} \in A$, that can be performed at the user's current station. For each time step in the game, the game interface displays the recommended next action; an example is shown in panel (d) of Figure \ref{fig:game_screen}.
The game interface also displays the game state, $S$, at any given time step in a user-friendly manner such as panel (c) in Figure \ref{fig:game_screen}. Specifically, $S = \{S_{l},S_{i}\}$ where $S_{l}$ defines the location of the chef and each ingredient (i.e which station), and $S_{i}$ defines the state of each ingredient,(i.e. chopped tomato, cooked chicken). Note, to mimic the complexities described in prior restaurant planning domains, we explicitly include the complexities of allowing parallelizable actions and tasks, variable action duration, and one-to-many mapping between actions and subgoals.

% Figure \ref{fig:system_diagram} provides an overview of the overall IDS system. When $a_{h}$ is performed, game state $S$ is updated and utilized by the underlying planner to provide a recommendation to the user, either in the form of a subgoal-explanation $\mathcal{E_{SB}}$, a causal-link-chain explanation $\mathcal{E_{CLC}}$ \cite{seegebarth2012making}, or as a default action recommendation $a_{IDS}$.
% Given a recommendation, the user can choice to conform to $a_{IDS}$ or select an alternative action (i.e. $a_{h} \neq a_{IDS}$). If the user selects an alternative action, the underlying planner in the game utilizes the updated state information to find a new plan solution $\pi^{'}$ from which subsequent actions are suggested to the user.

\subsection{Plan-Based IDS System}
\begin{figure*}
\centering
\includegraphics[width=15cm]{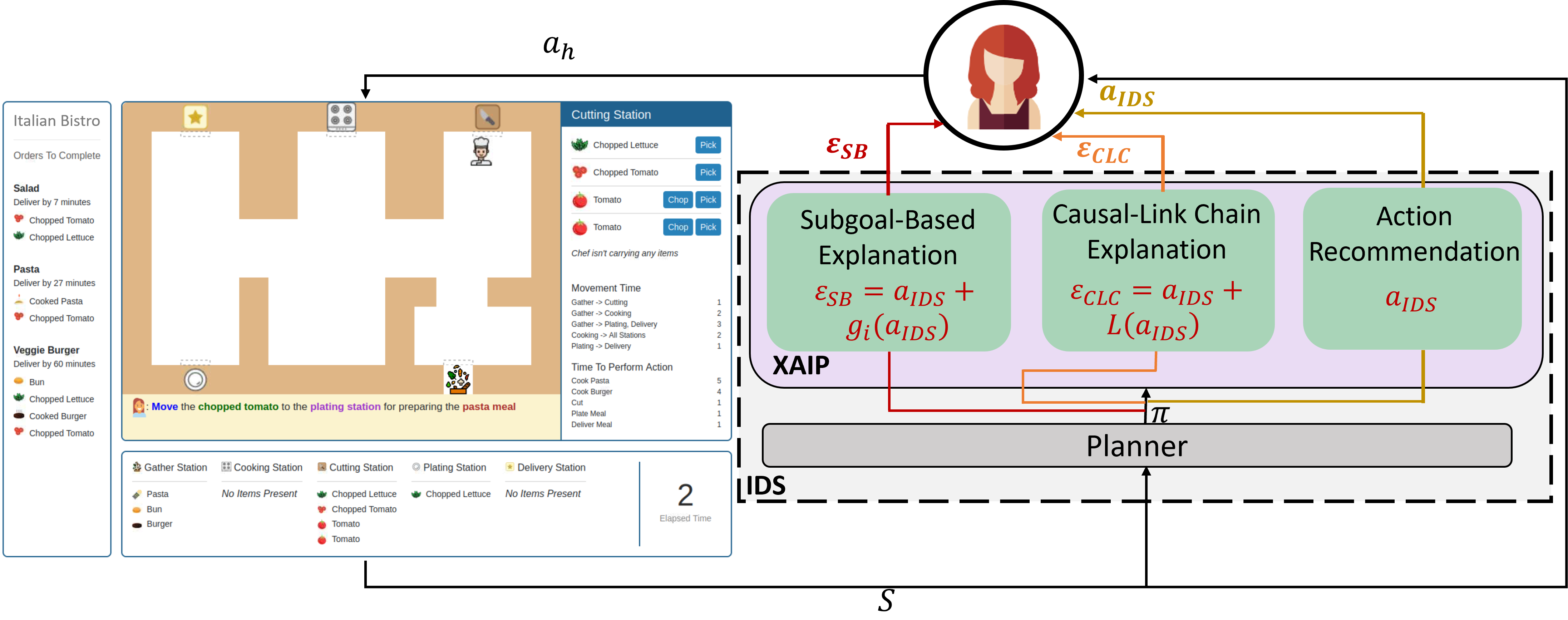}
\caption{Our IDS system utilized to evaluate the efficacy of $\mathcal{E_{SB}}$. A user performs any action $a_{h}$, which updates the environment $S$. The underlying planner utilizes $S$ to produce a plan solution $\pi$. The IDS system provides either  $\mathcal{E_{SB}}$, $\mathcal{E_{CLC}}$ or $a_{IDS}$. If the user ignores the planner's suggestion, the planner will replan for a new $\pi^{'}$ for subsequent action suggestions.}
\label{fig:system_diagram}
\end{figure*}

Figure \ref{fig:system_diagram} provides an overview of our overall IDS system. When a user action $a_{h}$ is performed, game state $S$ is updated and utilized by the underlying planner to find a plan solution $\pi$. Recall, in this scenario, solution $\pi$ consists of a sequence of actions, $\{a_0, a_1,..a_n\}$ that meets the task objective of delivering each required meal to restaurant customers within the goal meal delivery time, $t^m_{goal\_delivery} \forall m \in M$. The IDS system leverages $\pi$ to provide a recommendation to the user, either in the form of a subgoal-explanation $\mathcal{E_{SB}}$, a causal-link-chain explanation $\mathcal{E_{CLC}}$ \cite{seegebarth2012making}, or as a default action recommendation $a_{IDS}$. Given a recommendation, the user can choose either to conform to $a_{IDS}$ or select an alternative action (i.e. $a_{h} \neq a_{IDS}$). If the user selects an alternative action, the underlying planner in the game utilizes the updated state information to find a new plan solution $\pi^{'}$ from which subsequent actions are suggested to the user.

%\subsection{Restaurant IDS Planner}
To find a plan solution $\pi$, we represent the problem domain in PDDL \cite{fox2003pddl2}, a common programming language for planning problems, and utilize a state-of-the-art planner known as Temporal Fast Downward (TFD) \cite{eyerich2009using}. Although TFD is a temporal
planner with abilities to handle durative actions, we formulate
our planning problem as a classical planning problem in
which the objective is to minimize action costs as opposed
to duration.  We utilize TFD to more accurately model a restaurant domain where actions can occur simultaneously. To simulate multitasking, we leverage TFD's support for numeric fluents which allows us to track the cost of actions $a \in A$ being performed while an ingredient is cooking to ensure that an appropriate duration has elapsed for an ingredient to cook. 

To solve for a plan solution $\pi$, the planner utilizes the same action space $A$ and state space $S$ as that available to the user. The planner's initial state $\mathcal{I}$ is defined with select, pre-performed actions to guarantee that the planner can find a plan solution within 5 seconds for system usability, ensuring that suggestions and explanations are given to the user in a reasonable amount of time. The planner's goal state $\mathcal{G}$ is defined by reaching the \textit{delivered} state for all necessary $M$ meals. Each action $a \in A$ has a static cost of $c_{a}$ where the cost represents the time needed to perform $a$. If each meal $m$ is delivered at $t^{m}_{delivered}$, the objective of the planner is to minimize the overtime delivery cost, $\sum_{m=1}^{|M|} t^{m}_{delivered} -t^{m}_{goal\_delivery}$. In this work, the goal meal delivery times, $t^{m}_{goal\_delivery}$ are designed such that there exists one optimal solution for delivering all meals on time. \footnote{We leave the exploration of our IDS system under the context of partially ordered plans solutions, scenarios in which multiple, optimal solutions may exist, for future work.}

\subsection{Generating Suboptimal Plans}
A central objective of our work is to examine explanatory action suggestions in the context of \textit{suboptimal} IDS recommendations.  To achieve this objective, we intentionally corrupt the optimal plan, $\pi^*$, generated by our planner such that the resulting plan $\tilde{\pi}$ is suboptimal. At run time, we randomly select with probability $p$ whether a recommended action, $a_{IDS}$, is provided from an optimal or suboptimal plan. Specifically:
\[
    a_{IDS} =
\begin{cases}
    a_{IDS} \in \tilde{\pi},& \text{if } rand() \leq p\\
    a_{IDS} \in \pi^*,              & \text{otherwise}
\end{cases}
\]

Recall, the goal of the planner in our restaurant planning domain is to minimize the overtime in delivering each meal $m \in M$. Therefore, to generate a suboptimal plan $\tilde{\pi}$, we replace the optimal action of interest required for a particular meal $m_{i}$ with a random action required for some other random future meal $m_{j}$.  The resulting suboptimal action is therefore still relevant to the overall cooking task, and is not an obvious and trivially identifiable error (e.g., throw steak on floor). Reordering actions in this way is guaranteed to result in a suboptimal plan because it delays meals and leads meals to be completed out of order, causing the planner's overtime delivery cost to be non-zero.

% A central objective of our work is to examine explanatory action recommendations in the context of suboptimal IDS recommendations.  To achieve this objective, we
% intentionally corrupt the optimal plan generated by our planner, $\pi^*$, to generate a suboptimal plan $\tilde{\pi}$.  At run time, we randomly select with probability $p$ whether the recommended action provided is
% generated from the optimal or suboptimal plan.  
% Specifically:
% \[
%     a_{IDS} =
% \begin{cases}
%     \pi^*,& \text{if } rand() \leq p\\
%     \tilde{\pi},              & \text{otherwise}
% \end{cases}
% \]

% Recall, the goal of the planner in our restaurant planning domain is to minimize the overtime in delivering each meal $m \in M$. 
% Therefore, to generate suboptimal plan $\tilde{\pi}$, we randomly replace 12-14\% of optimal actions required to prepare a particular $m_{i} \in M$ with valid actions required to prepare an alternative meal $m_{j} \in M$. 
% In this manner, the planner's overtime cost for $\tilde{\pi}$ becomes non-zero, and the plan becomes suboptimal. 

\begin{center}
\begin{table*}
\begin{tabular}{ |p{2.5cm} |p{3.0cm}| p{4.1cm}| p{4.1cm}|}
\hline
 \textbf{Action ($a \in \pi$)} & \textbf{Action Recommendation ($a_{IDS}$)} & \textbf{Causal-Link-Chain Explanation ($\mathcal{E_{CLC}}$)} & \textbf{Subgoal-Based Explanation ($\mathcal{E_{SB}}$)}\\
\hline
 \textbf{cut} chef gatherStation cutStation tomato1 & Chop the tomato. & Chop the tomato for the salad meal or pasta meal or veggie burger meal. & Chop the tomato for the salad meal.  \\
 \hline
 \textbf{move-item} chef gatherStation plateStation salmon1 & Move the salmon to the cooking station. & Move the salmon to the cooking station to cook the salmon. &  Move the salmon to the cooking station for preparing the teriyaki salmon meal.\\
 \hline
 \textbf{end-cook} chef cookStation broth1  &  Finish cooking the broth. & Finish cooking the broth for preparing the soup. & Finish cooking the broth for preparing the soup. \\
 \hline
  \textbf{move-chef} chef plateStation gatherStation  & Move to the gather station. & Move to the gather station to move the tomato. & Move to the gather station for preparing the pasta meal.\\
\hline
\end{tabular}
\caption{Action recommendations ($a_{IDS}$), causal-link chain explanations $\mathcal{E_{CLC}}$ and subgoal explanations ($\mathcal{E_{SB}}$) for select action $a \in \pi$. Note, $a$ is represented in PDDL \cite{fox2003pddl2}, a standard language definition for programming planning problems; however for legibility, the key action is emphasized in bold.}
\label{tab:Exps}
\end{table*}
\end{center}

\subsection{Generating Subgoal-Based Explanations}
Given a set of subgoals $\langle g_{0}...g_{n}\rangle$ (defined a priori), we employ a post-hoc search to map actions $\langle a_{0}...a_{n}\rangle$ within $\pi$ with a corresponding subgoal $g_{i} \in G$.  In our work, subgoals are defined as the designated meal for which an action $a$ is being performed. To present $\mathcal{E_{SB}}$ in a manner understandable by novice users, we leverage templated, natural language. We parse each action $a$ output in $\pi$ to retrieve the contextual information that action $a$ is acting upon. In our work, the contextual information corresponds to the \textit{ingredient(s)} the action would be applied on or the \textit{location} the action would be applied to. In this manner, we template our explanation as follows, \textit{``$\langle$action$\rangle$ the $\langle$ingredients \slash location$\rangle$ for $\langle g_{i}(a)\rangle$"}. 

We compare our subgoal explanations, $\mathcal{E_{SB}}$, to notable baselines from prior work: action suggestions, $a_{IDS}$, and causal link chain explanations \cite{seegebarth2012making}, $\mathcal{E_{CLC}}$, from the XAIP community. A causal link $l$ is defined as (${s\rightarrow}_{p}$ $s'$) where $p$ denotes the set of preconditions for future plan step(s) $s'$ that are met by the effects of the current plan step $s$ \cite{seegebarth2012making}. Thus $\mathcal{E_{CLC}}$ justify the action $a$ in the current plan step by indicating the set of satisfied causal links $l \in L$. Alternatively, $a_{IDS}$ models the default output of current IDS systems. 

% In Table \ref{tab:Exps}, we provide example explanations of our $\mathcal{E_{SB}}$ explanations in comparison to notable baselines: action suggestions, $a_{IDS}$,  and causal link chain explanations \cite{seegebarth2012making}, $\mathcal{E_{CLC}}$, from the XAIP community. A causal link $l$ is defined as (${s\rightarrow}_{p}$ $s'$) where $p$ denotes the set of preconditions for future plan step(s) $s'$ that are met by the effects of the current plan step $s$ \cite{seegebarth2012making}. Thus $\mathcal{E_{CLC}}$ justify the action $a$ in the current plan step by indicating the set of satisfied causal links $l \in L$. Alternatively, $a_{IDS}$ models the default output of current IDS systems. 

In Table \ref{tab:Exps}, we provide examples highlighting differences between our $\mathcal{E_{SB}}$ explanations, causal link chain explanations \cite{seegebarth2012making}, and action suggestions. For example, the first entry in Table \ref{tab:Exps} indicates a level of ambiguity causal-link-chain explanations may present since they enumerate \textit{all} preconditions a current action meets. In contrast, our subgoal explanations provide only the most appropriate subgoal related to a given action. Additionally, as seen by all the examples in Table \ref{tab:Exps}, causal link chain explanations do not differentiate between the \textit{type} of precondition being fulfilled. In some instances the preconditions are other primitive actions (second and third example), and in other instances when there is only one action required for a subgoal, the met precondition is the subgoal. Therefore, causal-link- explanations \textit{can} be identical to subgoal-explanations, such as the third example in Table \ref{tab:Exps}; however, this occurrence is rare in complex tasks since it is an artifact of a subgoal requiring only a single action for completion. In contrast, our subgoal-explanations \textit{always} highlight the relevant subgoal for a given action, instead of focusing on action-level preconditions, providing a more consistent level of abstraction for complex tasks.

%Another type of explanation from the XAIP community that would be of future interest to investigate is the role of contrastive-styled explanations (CE) with subgoal information. In this work, we do not explicitly benchmark against CEs because of several differences between our research objectives and how CEs are used in prior work. Specifically, prior CE work assumes the planner is optimal, whereas we specifically don’t assume optimality. Additionally, CE relies on a user’s foil to be present (either queried or inferred); the foil is not available in our framework.

\section{Study Design}
Our primary goal is to evaluate the effect $\mathcal{E_{SB}}$ explanations have on user task performance in IDS settings. Recall, that in real world settings, IDS systems may be non-robust, sometimes providing suboptimal suggestions to end-users \cite{guerlain2000intelligent}. To examine the utility of $\mathcal{E_{SB}}$ in both suboptimal and optimal IDS settings,
we conducted a seven-way, between-subjects study in which participants were asked to play the restaurant planning game. 
Specifically, we used a 3 x 2 factorial study design with one factor being the type of IDS (with $a_{IDS}$ suggestions, $\mathcal{E_{CLC}}$ or $\mathcal{E_{SB}}$ explanations) and the second factor being optimality of IDS recommendation (optimal IDS and suboptimal IDS).  The seventh study condition was an additional baseline study condition in which participants did not receive any help from an IDS. Below, we detail each study condition:
\begin{itemize}
    \item {\underline{None (Baseline)}}: Participants receive no suggestions from an IDS system.
    \item {\underline{$\pi(a_{IDS})$ (Baseline)}}: Participants receive action recommendations from an optimal IDS system, and is closely modeled after the default action suggestion features available in plan-based IDS systems \cite{grover2020radar}.
    \item {\underline{$\pi(\mathcal{E_{CLC}})$ (Baseline)}}: Participants receive causal link chain explanations from an optimal IDS system and is closely modelled after the causal link explanations in \cite{seegebarth2012making}.
    \item {\underline{$\pi(\mathcal{E_{SB}})$}}: Participants receive subgoal-based explanations from an optimal IDS system.
    \item {\underline{$\tilde{\pi}(a_{IDS})$}}: Participants receive action recommendations from a suboptimal IDS system.
    \item {\underline{$\tilde{\pi}(\mathcal{E_{CLC}})$}}: Participants receive causal link explanations from a suboptimal IDS system.
    \item {\underline{$\tilde{\pi}(\mathcal{E_{SB}})$}}: Participants receive subgoal-based explanations from a suboptimal IDS system.
\end{itemize}

The study consisted of three stages in which participants played a total of five games, each consisting of a unique set of $M$ meals to prepare. Participants proceeded to the next game when they finished delivering all required $M$ meals, or when time cost in a game reached 80, whichever came first. The study consisted of three stages: the familiarization stage, IDS stage and an assessment stage, detailed below.

\textit{\underline{Familiarization:}} The participants first played through an interactive tutorial which explained the components of the interface as the participants made a burrito meal. While the interactive tutorial remained the same across all study conditions, the \textit{None} study condition did not receive support from the IDS system, whereas all other study conditions received their respective guidance from the IDS system. Participants also played a second game to get further acquainted with the system. In the practice round, participants were tasked with making two meals (BLT sandwich, hotdog), and participants received IDS based on their study conditions. The familiarization stage was designed to familiarize users with all aspects of the interface and to minimize learning effects in future games.

\textit{\underline{IDS:}} Participants played two more games, each with the objective of preparing three meals with the help from an IDS system (or no help in the \textit{None} study condition). These games were themed by cuisine: Italian Bistro (salad, pasta, veggie burger) and Asian Fusion (chicken quesadilla, soup, sushi). Prior to playing, participants were told that the anthropomorphized IDS system, Manager Molly, may provide suboptimal suggestions. The goals of the participants in all study conditions were to delivery meals on time and to identify suboptimal suggestions. Both games were counterbalanced, such that a random 50\% of participants played the Italian Bistro game first, while remaining played the Asian Fusion game first. In the two suboptimal recommendation study conditions ($\tilde{\pi}(a_{IDS})$ and $\tilde{\pi}(\mathcal{E_{SB}})$), $p=0.85$, such that the accuracy of our IDS was 85\%, and approximately 15\% of recommendations viewed by the users were corrupted to be suboptimal \footnote{The level of acceptable error in a deployable IDS system varies significantly by application (e.g., medical diagnosis systems may be expected to perform with greater accuracy than those in lower risk domains).  From prior work, we find  accuracy rates of 75-95\% across applications \cite{rodriguez2020intelligent,RATHORE2018920}. We selected an accuracy of 85\% as it models many state of the art systems.}.

\textit{\underline{Assessment:}} Participants in all seven study conditions played a final game, with \textbf{no support from the IDS system}. Similar to the IDS stage, the assessment game required delivering 3 meals (teriyaki salmon, steak \& potatoes, chili). In this assessment,  participants in all study conditions had one objective, which was to deliver meals by their designated delivery time. The goal of the assessment stage was to simulate a scenario where a failure occurs in an IDS system, and guidance is no longer available to a novice user. Our goal is to understand how previous exposure to IDS in an optimal or suboptimal setting may impact participant performance on a task when in the absence of any IDS.

%($a_{IDS}$ or $\mathcal{E_{SB}})$, may impact participant performance on a task in the absence of any IDS. 

\subsection{Metrics}
To measure user task performance and overall understanding of the task, we evaluate three metrics: 
\begin{itemize}
    \item \textbf{User Plan Cost} (\textit{UPC}): represents participant overtime cost in delivering meals \textit{per game}. This metric is analogous to how the planner optimizes for the optimal plan solution. Below, $M$ represents the total number of meals to complete for a game, $t^{m}_{delivered}$ represents the accumulated time cost at which meal $m$ is delivered, and $t^{m}_{goal\_delivered}$ represents the designated time cost at which a meal should have been delivered.
    \begin{equation}
        UPC = \sum_{m=1}^{|M|} t^{m}_{delivered} -t^{m}_{goal\_delivery} 
    \end{equation}
    \item \textbf{Optimal Action Conformance} (OAC\%): represents the total percentage of optimal actions suggested from the IDS system that participants performed.
    \item \textbf{Suboptimal Action Avoidance} (SAA\%): represents the total percentage of suboptimal actions suggested from the IDS system that participants avoided.
\end{itemize}
Both \textit{OAC} and \textit{SAA} are measured during the \textit{IDS Stage} of the user study, while \textit{UPC} is measured for games within the \textit{IDS Stage} and \textit{Assessment Stage}.

Additionally, we utilize the following metric to understand user preferences in IDS system outputs: 
\begin{itemize}
    \item \textbf{Perceived Preference} (\textit{Pref\%}): represents the total percentages of $a_{IDS}$, $\mathcal{E_{CLC}}$ or $\mathcal{E_{SB}}$ IDS output types preferred by participants for understanding the chef's next action.
\end{itemize}

%Given these metrics, our data analysis seeks to answer the following questions: 
%\begin{itemize}
%    \item \textbf{Q1}: In the presence of an IDS system, do SB explanations improve users' task performance? 
    % Q1. what type of perfect/imperfect explanations help improve users' task performance at training time compare to no explanation at all?
    % A1. both types, perfect and imperfect helps at training time. Figure 2.
    % Q2. what type of imperfect explanation help users to detect that they are flawed?
    % A2. At least in one scenario, SB seems to help. Figure 3.
    % Q3. what type of perfect/imperfect explanations help at deployment time? 
    % A3. only SB helps even if the explanations weren't perfect at training time.  Figure 4.
%    \item \textbf{Q2}: Do SB explanations help users distinguish incorrect and correct advice in imperfect IDS systems?
%    \item \textbf{Q3}: Can SB explanations aid users in developing a deeper understanding of the underlying task such that in the absence of an IDS, users are able to have high task performance?
%\end{itemize}

\subsection{Participants}
We recruited 120 individuals from Amazon's Mechanical Turk. We filtered participants that showed no effort to play the game in the form of taking repeated actions until the timeout. The filtration process yielded 105 remaining participants (53 males, 52 females). All participants were 18 years or older (M = 36.3, SD =10.3). Each study condition had 15 participants. The task took on average 40 minutes and participants were compensated \$5.00.

\section{Study Results}
All metrics,  Optimal Action Conformance (\textit{OAC}), Suboptimal Action Avoidance (\textit{SAA}) User Plan Cost (\textit{UPC}), and Perceived Preference (\textit{Pref\%}) were analyzed with a one-way ANOVA, followed by a post-hoc Tukey Test. 

\begin{figure*}[!t]
\centering
\begin{subfigure}[b]{0.4\textwidth}
  \centering
  \includegraphics[width=\textwidth]{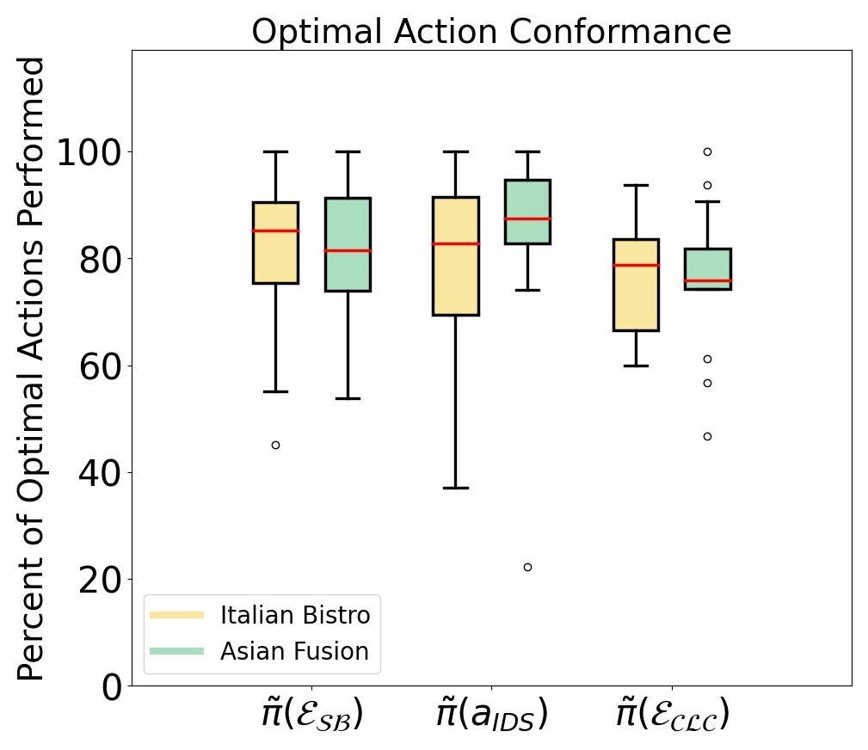}
  \caption[]%
  {{}}
\end{subfigure}\quad
\begin{subfigure}[b]{0.4\textwidth}
  \centering
  \includegraphics[width=\textwidth]{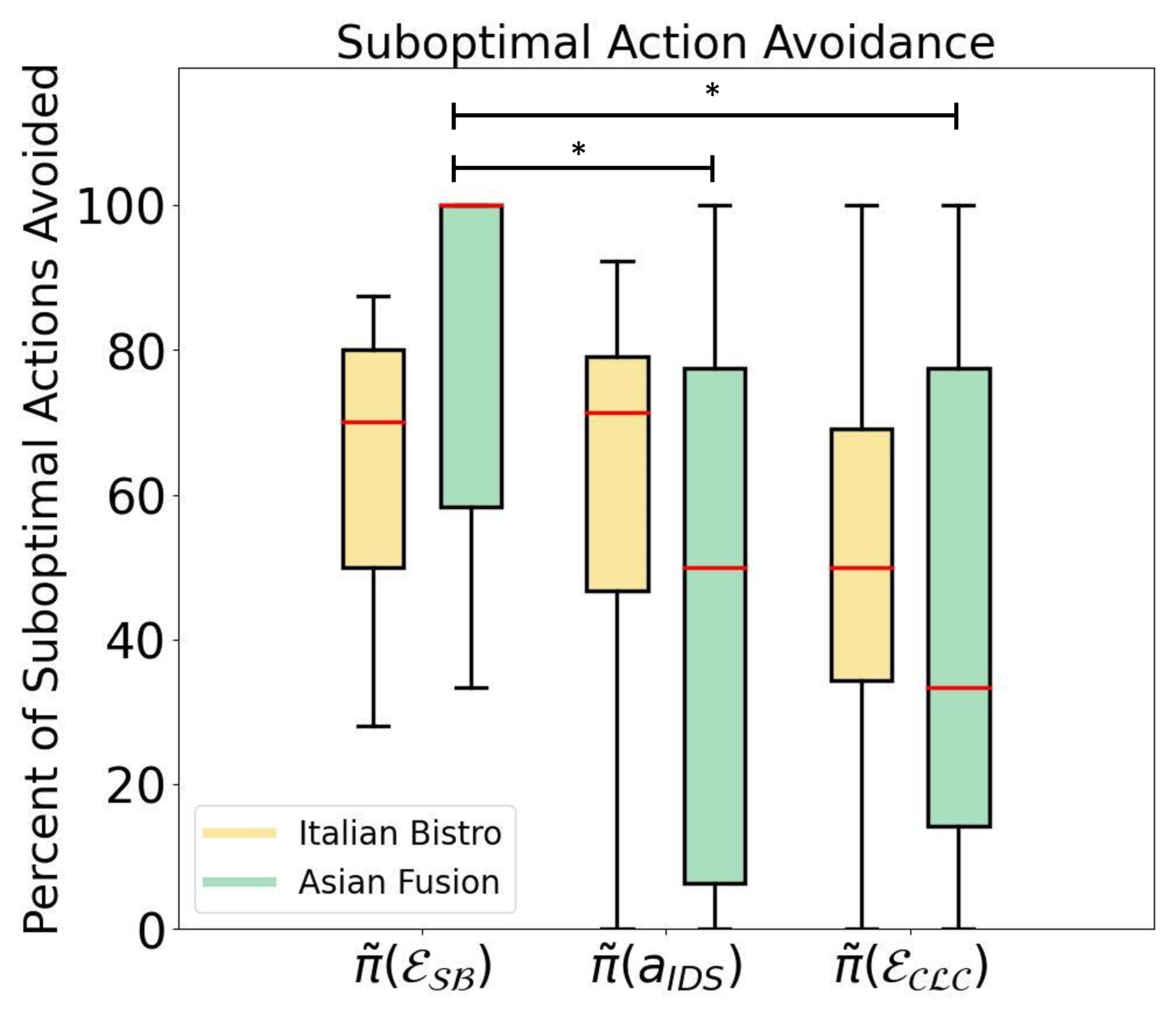}
    \caption[]%
     {{}}
     \end{subfigure}
\caption{User optimal action conformance and action avoidance percentages for participants that received $\mathcal{E_{SB}}$, $\mathcal{E_{CLC}}$ and $a_{IDS}$, from suboptimal IDS systems. Statistical significance is reported as: *p \textless 0.05.}

\label{fig:optimality}
\end{figure*}

\textbf{H1:} In Figure \ref{fig:optimality}, we present user optimal action conformance (\textit{OAC\%}) as well as suboptimal action avoidance (\textit{SAA\%}) in order to analyze the benefit of providing subgoal-based explanations, $\mathcal{E_{SB}}$, for understanding suboptimality in IDS systems. We observe in Figure \ref{fig:optimality}(a) that $\tilde\pi(a_{IDS})$, $\tilde\pi(\mathcal{E_{CLC}})$ and $\tilde\pi(\mathcal{E_{SB}})$ study conditions had similarly high \textit{OAC\%}s. In other words, action recommendations, causal link chain explanations and subgoal-based explanations helped participants discern optimal recommendations. However, in Figure \ref{fig:optimality}(b), we observe that participants in the $\tilde\pi(a_{IDS})$ study condition and  $\tilde\pi(\mathcal{E_{CLC}})$ study condition had much lower \textit{SAA\%s} than participants in the $\tilde\pi(\mathcal{E_{SB}})$ study condition. In fact, in the Asian Fusion game, we observe a significant difference in \textit{SAA\%} between the $\tilde\pi(a_{IDS})$ and $\tilde\pi(\mathcal{E_{SB}})$ study conditions (t(42)=-2.80, $p<0.05$) as well as between the $\tilde\pi(\mathcal{E_{CLC}})$ and $\tilde\pi(\mathcal{E_{SB}})$ study conditions (t(42)=-2.95, $p<0.05$). \textbf{These results support H1}, indicating that in the context of an suboptimal IDS system, subgoal-based explanations, $\mathcal{E_{SB}}$, help participants detect and avoid more suboptimal suggestions compared to those who only receive $a_{IDS}$ or $\mathcal{E_{CLC}}$.

\begin{figure*}
\centering
\begin{subfigure}[b]{0.4\textwidth}
  \centering
  \includegraphics[width=\textwidth]{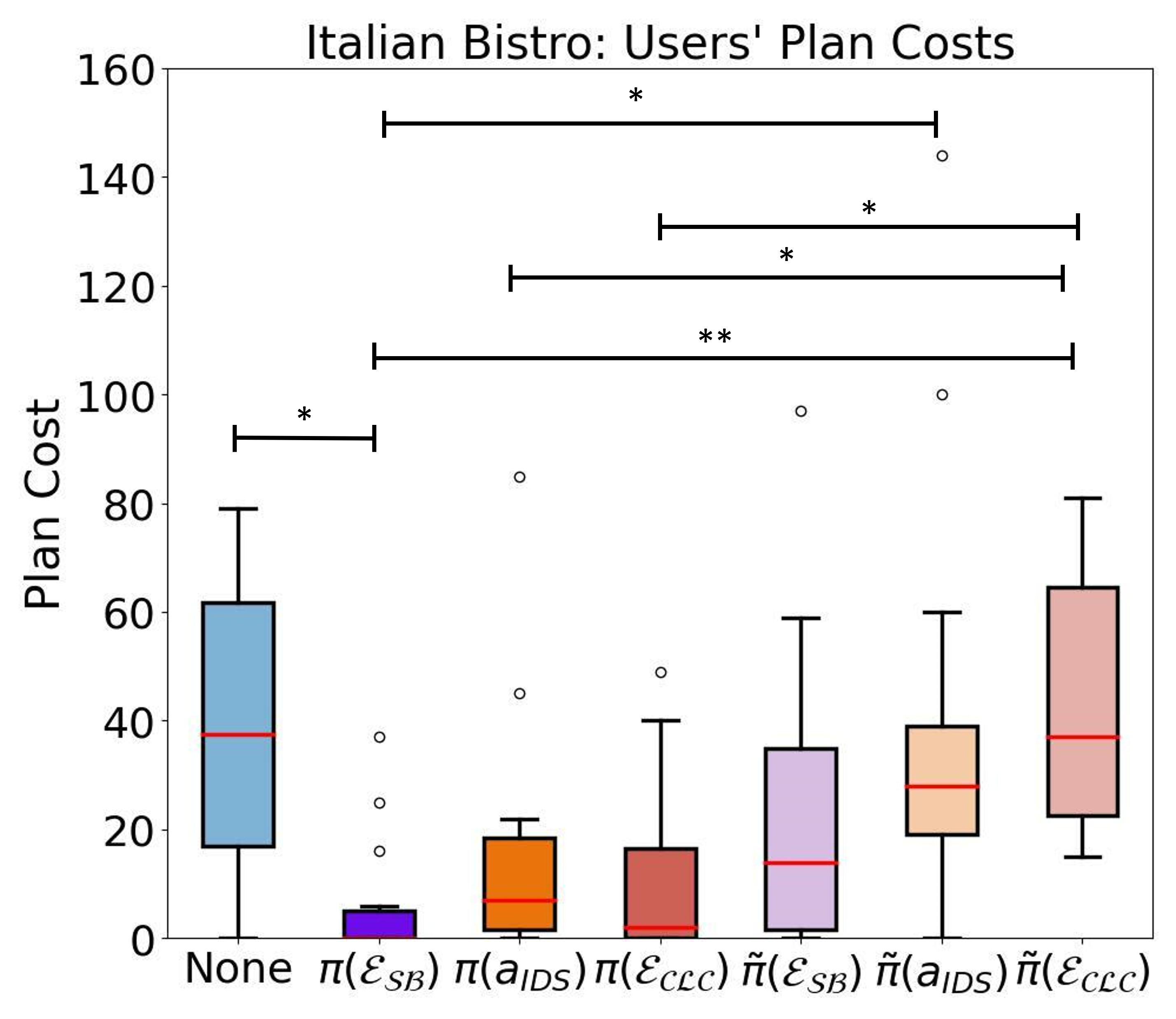}
  \caption[]%
  {{}}
\end{subfigure}\quad
\begin{subfigure}[b]{0.4\textwidth}
  \centering
  \includegraphics[width=\textwidth]{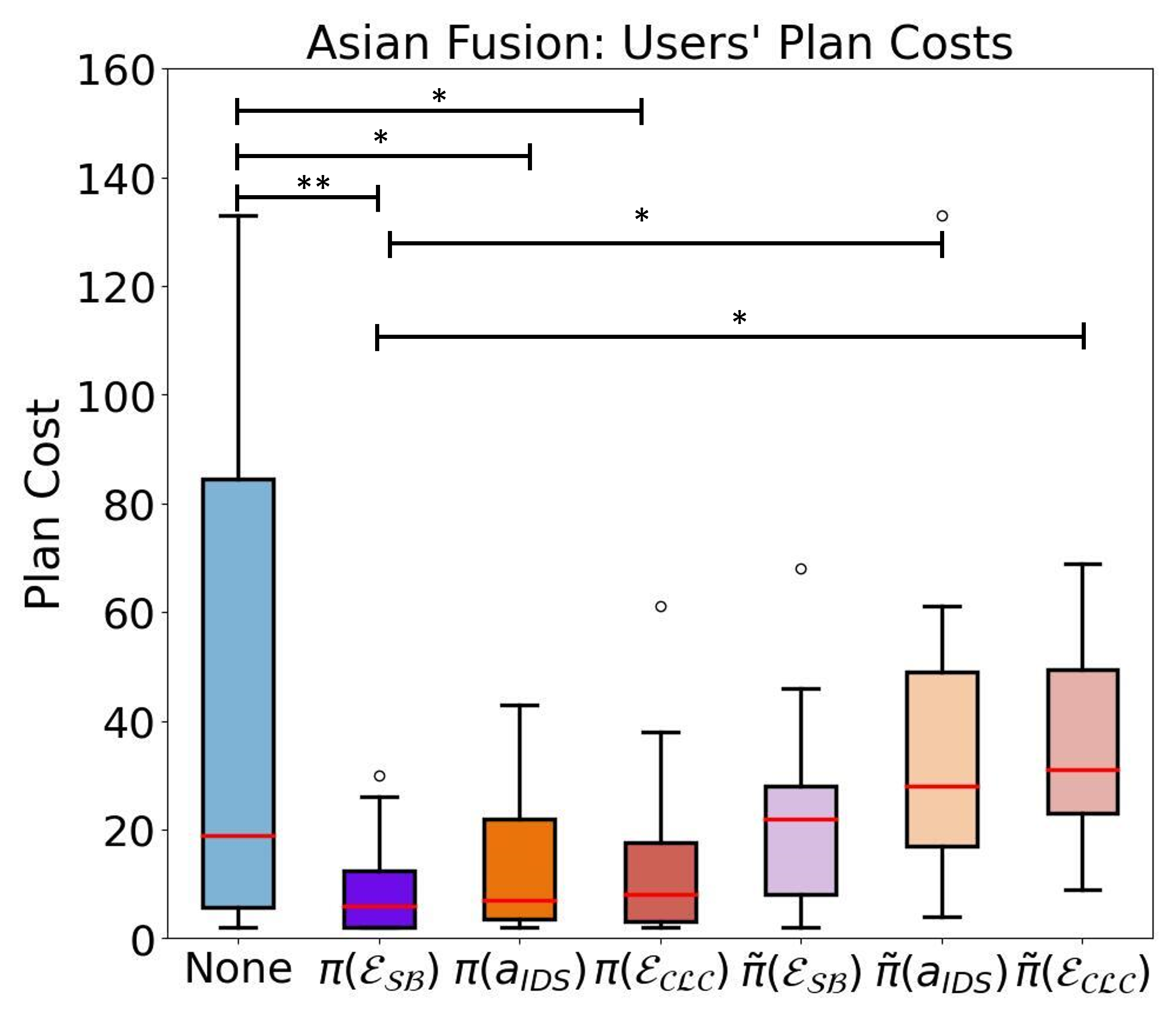}
    \caption[]%
     {{}}
     \end{subfigure}
\caption{User Plan Cost across all study conditions for the two themed games within the \textit{IDS Stage} of the user study. Statistical significance is reported as: *p \textless 0.05, **p \textless 0.01.}
\label{fig:plancosts}
\end{figure*}

\textbf{H3:} In Figure \ref{fig:plancosts}, we present user plan costs (\textit{UPC}) across each study condition in the two themed games to analyze the impact of including subgoal information on IDS-supported user performance. Overall, we observe that in both games participants in the  $\pi(\mathcal{E_{SB}})$ study condition, subgoal-based explanations from an optimal IDS system, have the best overall task performance in comparison to the other study conditions. Additionally, out of the explanation-based study conditions, we observe $\mathcal{E_{CLC}}$ explanations to lead to the highest plan cost under suboptimal IDS across both games. Participants in $\pi(\mathcal{E_{SB}})$ had significantly lower \textit{UPC} compared to participants in the \textit{None} study condition for both the Italian Bistro game (t(95)=-3.46, $p<0.05$) and the Asian Fusion game (t(101)=-4.00, $p<0.01$). We additionally observe that participants in the $\pi(\mathcal{E_{SB}})$ study condition had significantly lower \textit{UPC} than participants in the $\tilde\pi(a_{IDS})$ study condition for the Asian Fusion game (t(101)=8.17, $p<0.05$). Moreover,  we observe that participants in $\pi(\mathcal{E_{SB}})$ had significantly lower \textit{UPC} compared to participants in the $\tilde\pi(\mathcal{E_{CLC}})$ study condition in both the Italian Bistro (t(95)=4.08, $p<0.01$) and Asian Fusion game (t(101)=3.10, $p<0.05$). \textbf{These results support H3} by indicating that supplementing IDS outputs with explanations grounded in subgoal information help participants understand the underlying motivation for a suggestion, and therefore perform the task significantly better those who only receive $a_{IDS}$ or even causal-link based explanations ($\mathcal{E_{CLC}}$).

\begin{figure}[!t]
    \centering
    \includegraphics[width=0.4\textwidth]{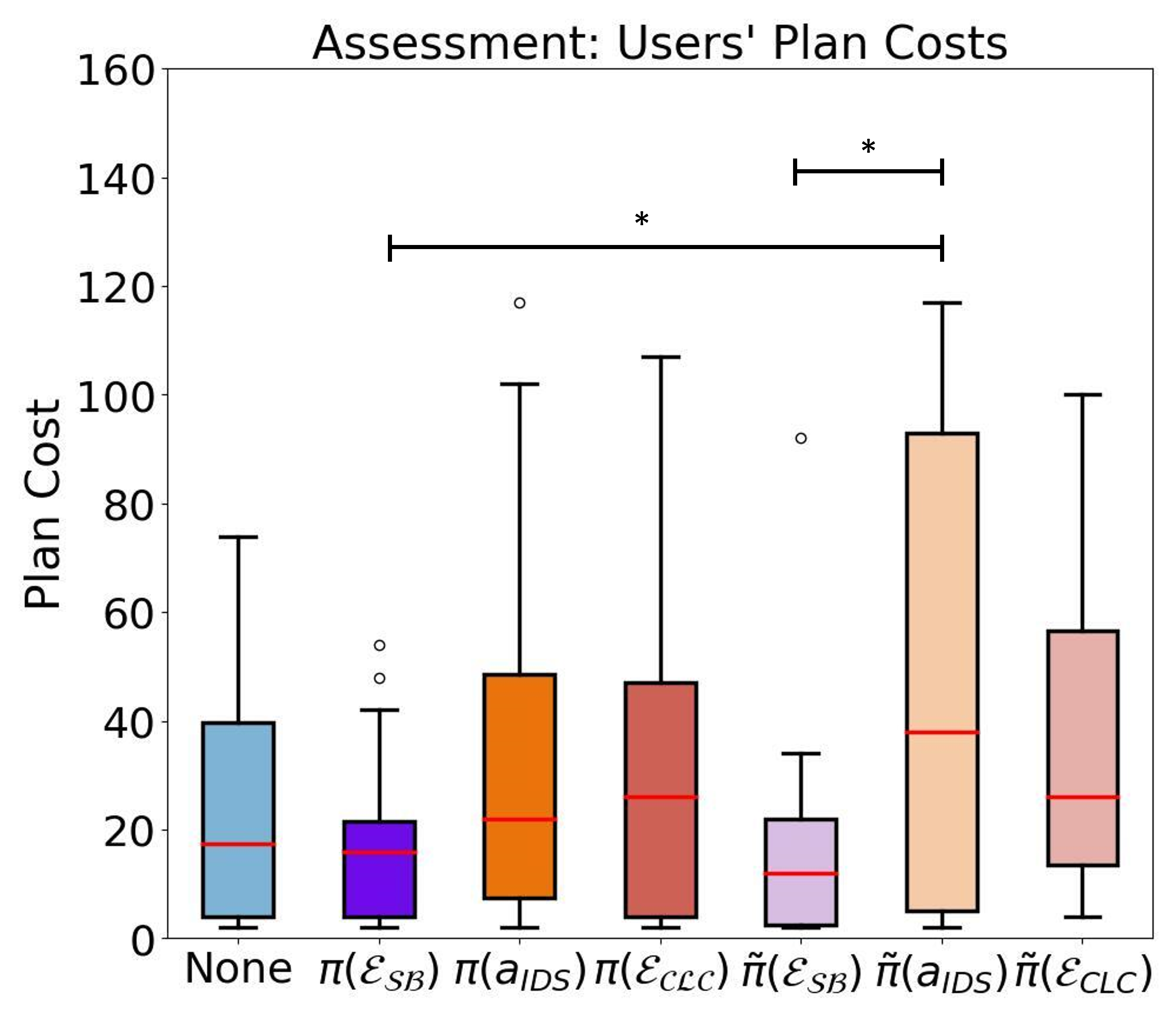}
    \caption{User Plan Costs for the assessment game in which all participants did not receive any guidance from an IDS system. Statistical significance is reported as: *p \textless 0.05.}
    \label{fig:assessment}
\end{figure}

\textbf{H2:} In Figure \ref{fig:assessment}, we present user plan cost (\textit{UPC}) from the Assessment stage across each study condition. None of the participants had access to IDS recommendations in this game, and the results allow us to assess how prior exposure to $\mathcal{E_{SB}}$ impacts user performance once IDS recommendations are unavailable. Overall, we observe that both $\pi(\mathcal{E_{SB}})$ and $\tilde\pi(\mathcal{E_{SB}})$ study conditions have the lowest \textit{UPC} compared to the other study conditions, including prior work's $\mathcal{E_{CLC}}$ explanations. In fact, participants in the $\pi(\mathcal{E_{SB}})$ study condition, those previously received $\mathcal{E_{SB}}$ explanations under an optimal IDS system, have significantly lower \textit{UPC} than participants in the $\tilde\pi(a_{IDS})$ study condition (t(67)=2.83, $p<0.05$). Similarly, we observe participants in the $\tilde\pi(\mathcal{E_{SB}})$ study condition, those who previously received $\mathcal{E_{SB}}$ explanations under a suboptimal IDS system, also have significantly lower \textit{UPC} than those who received $\tilde\pi(a_{IDS})$ (t(67)=2.87, $p<0.05$). \textbf{These results support H2}, demonstrating the important role of subgoal-based explanations, $\mathcal{E_{SB}}$, in training users to understand the underlying task, compared to action-based recommendations $a_{IDS}$, even if IDS was suboptimal during training.

\begin{figure}[!t]
    \centering
    \includegraphics[width=0.45\textwidth]{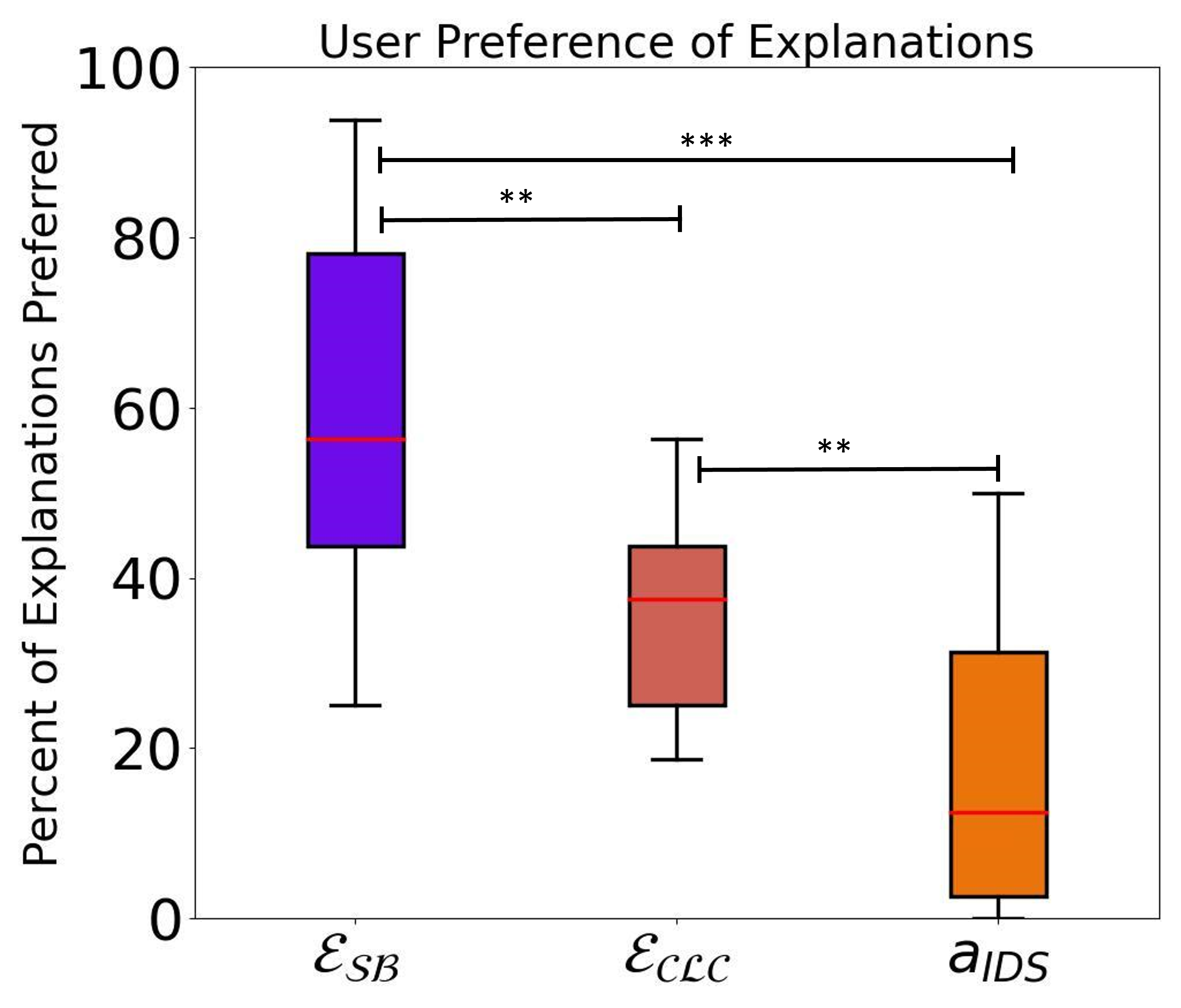}
    \caption{User perceived preferences towards $a_{IDS}$, $\mathcal{E_{CLC}}$ and $\mathcal{E_{SB}}$ in understanding the chef's next action. Statistical significance is reported as: **p \textless 0.01, ***p \textless 0.001.}
    \label{fig:preference}
\end{figure}

\textbf{H4:} To evaluate H4 we conducted an additional experiment in which we presented users with three IDS system output options, $a_{IDS}$, $\mathcal{E_{CLC}}$, and $\mathcal{E_{SB}}$, and asked them to select their preferred IDS system output\footnote{Since our previous study was between-subjects and participants were only exposed to one type of IDS, $\mathcal{E_{SB}}$, $\mathcal{E_{CLC}}$, or $a_{IDS}$, we conducted an additional within-subjects study with 20 participants from AMT (Male=11, Female=9, mean=36.4 SD=8.7) to measure user preference between the three types of IDS system outputs.}. Specifically, each participant was presented with 25 randomly shuffled, pre-recorded videos of optimal actions the chef would perform while preparing meals in the restaurant game. Each video was 10-15 seconds in duration and included two actions that the chef performed towards the goal. Participants were tasked with watching each video, to gain contextual understanding of which portion of the task the chef was working on, and select which form of the provided IDS output, $\mathcal{E_{SB}}$, $\mathcal{E_{CLC}}$ or $a_{IDS}$, they preferred in understanding the chef's next action. Figure \ref{fig:preference} presents the Perceived Preference (\textit{Pref\%}) metric results for the above experiment, which was analyzed with a one-way ANOVA with post-hoc Tukey Test. We observe that participants significantly prefer $\mathcal{E_{SB}}$ explanations compared to both $\mathcal{E_{CLC}}$ explanations (t(54)=-3.50, $p<0.01$), and $a_{IDS}$ (t(54)=-6.66, $p<0.001$). Additionally, we observe that $\mathcal{E_{CLC}}$ explanations are significantly preferred compared to $a_{IDS}$ (t(54)=-3.17, $p<0.01$). \textbf{These results support H4} demonstrating that $\mathcal{E_{SB}}$ explanations are more frequently preferred by users, a factor that may aid in adoption of subgoal-based explanations for IDS systems.

\section{Discussion \& Conclusion}
In this work, we relax a critical assumption of prior explainable AI research related to plan-based IDS, and consider a scenario where IDS systems can sometimes be non-robust and unreliable, occasionally recommending a suboptimal action or becoming unavailable after a period of time. Leveraging insights from psychology, that humans naturally break complex tasks into smaller subgoals, we introduce a new, subgoal-based explanation type,  $\mathcal{E_{SB}}$, for plan-based IDS systems that explains a recommended action by grounding the action in the resulting subgoal that is satisfied. We validate the usefulness of $\mathcal{E_{SB}}$ through a user study involving a complex, restaurant planning game in which users play as a chef to prepare and deliver a set number of meals within each meal's delivery time. Our user study findings support our hypotheses, demonstrating that subgoal-based explanations ($\mathcal{E}_{SB}$) improve user task performance, improve user ability to distinguish optimal and suboptimal IDS recommendations, are preferred by users, and enable more robust user performance in  case of IDS failure. We find the results from the Assessment stage of the study (H2) particularly surprising.  All users received detailed task instructions and performed the same tutorials; the Assessment game was the 5th in the series of games, meaning that participants were reasonably familiar with the task at this stage.  Yet $\mathcal{E}_{SB}$ significantly impacted performance such that in both the optimal and suboptimal IDS study conditions, users learned the task objectives better than users who received causal-link explanations, $\mathcal{E}_{CLC}$, and those who received only action recommendations $a_{IDS}$. These results point to important benefits explanation-based IDS systems can have in real-world deployments, highlighting that even suboptimal IDS systems can serve as a useful training tool for users when $\mathcal{E}_{SB}$ are added to IDS output. To our knowledge, this is the first investigation of $\mathcal{E}_{SB}$ in suboptimal IDS systems.

While these results are promising, our work has several limitations that present opportunities for future work.  First, we conducted our study only with novice users\footnote{Our domain differs from both real world cooking and the Overcooked online game, and thus has significant novelty to all users.}.  Further studies should explore whether the observed benefits of $\mathcal{E}_{SB}$ hold for expert users. Second, we conducted our study over a limited period of time, and thus factors such as long-term learning effects, fatigue, and automation bias were not fully explored.  Further work is needed to fully explore the effect that explanations have on long-term IDS deployment.  Third, our subgoals were predefined. Coupling our approach with existing techniques that autonomously identify subgoals may yield new insights on the usability of $\mathcal{E}_{SB}$ in plan-based IDS systems. Furthermore, we validate the utility of $\mathcal{E}_{SB}$ with an acceptable probability of error in IDS systems noted in prior work. However, future investigation is needed to examine other probability of errors in which benefits of $\mathcal{E}_{SB}$ continue to exist or when the benefits of $\mathcal{E}_{SB}$ may diminish. Additionally, given the usefulness of contrastive explanations in many XAIP applications where users may prompt back-and-forth ``Why X and not P?" questions to an intelligent system, it is important to examine the utility of subgoal-based explanations in a contrastive question-answering setting. Finally, further investigation is needed to see the benefits of $\mathcal{E}_{SB}$ when subgoals are more hierarchical. For example, in complex tasks there may exist multiple goals where goals have interdependencies. In these scenarios, there may also be multiple optimal solutions. An open ended question on how to retrieve meaningful subgoals in these complex hierarchical settings with partially ordered plan solutions still remains.

\bibliographystyle{ACM-Reference-Format}
\bibliography{references}


\begin{thebibliography}{47}

%%% the Web address.

\ifx \showCODEN    \undefined \def \showCODEN     #1{\unskip}     \fi
\ifx \showDOI      \undefined \def \showDOI       #1{#1}\fi
\ifx \showISBNx    \undefined \def \showISBNx     #1{\unskip}     \fi
\ifx \showISBNxiii \undefined \def \showISBNxiii  #1{\unskip}     \fi
\ifx \showISSN     \undefined \def \showISSN      #1{\unskip}     \fi
\ifx \showLCCN     \undefined \def \showLCCN      #1{\unskip}     \fi
\ifx \shownote     \undefined \def \shownote      #1{#1}          \fi
\ifx \showarticletitle \undefined \def \showarticletitle #1{#1}   \fi
\ifx \showURL      \undefined \def \showURL       {\relax}        \fi
% invisible to TeX
\providecommand\bibfield[2]{#2}
\providecommand\bibinfo[2]{#2}
\providecommand\natexlab[1]{#1}
\providecommand\showeprint[2][]{arXiv:#2}

\bibitem[Adadi and Berrada(2018)]%
        {adadi2018peeking}
\bibfield{author}{\bibinfo{person}{Amina Adadi} {and} \bibinfo{person}{Mohammed
  Berrada}.} \bibinfo{year}{2018}\natexlab{}.
\newblock \showarticletitle{Peeking inside the black-box: a survey on
  explainable artificial intelligence (XAI)}.
\newblock \bibinfo{journal}{\emph{IEEE access}}  \bibinfo{volume}{6}
  (\bibinfo{year}{2018}), \bibinfo{pages}{52138--52160}.
\newblock


\bibitem[Adebayo et~al\mbox{.}(2020)]%
        {adebayo2020debugging}
\bibfield{author}{\bibinfo{person}{Julius Adebayo}, \bibinfo{person}{Michael
  Muelly}, \bibinfo{person}{Ilaria Liccardi}, {and} \bibinfo{person}{Been
  Kim}.} \bibinfo{year}{2020}\natexlab{}.
\newblock \showarticletitle{Debugging tests for model explanations}.
\newblock \bibinfo{journal}{\emph{arXiv preprint arXiv:2011.05429}}
  (\bibinfo{year}{2020}).
\newblock


\bibitem[Arnold et~al\mbox{.}(2004)]%
        {arnold2004impact}
\bibfield{author}{\bibinfo{person}{Vicky Arnold}, \bibinfo{person}{Philip~A
  Collier}, \bibinfo{person}{Stewart~A Leech}, {and} \bibinfo{person}{Steve~G
  Sutton}.} \bibinfo{year}{2004}\natexlab{}.
\newblock \showarticletitle{Impact of intelligent decision aids on expert and
  novice decision-makers’ judgments}.
\newblock \bibinfo{journal}{\emph{Accounting \& Finance}} \bibinfo{volume}{44},
  \bibinfo{number}{1} (\bibinfo{year}{2004}), \bibinfo{pages}{1--26}.
\newblock


\bibitem[Canal et~al\mbox{.}(2021)]%
        {canal2021task}
\bibfield{author}{\bibinfo{person}{Gerard Canal}, \bibinfo{person}{Senka
  Krivic}, \bibinfo{person}{Paul Luff}, {and} \bibinfo{person}{Andrew Coles}.}
  \bibinfo{year}{2021}\natexlab{}.
\newblock \showarticletitle{Task Plan verbalizations with causal
  justifications}. In \bibinfo{booktitle}{\emph{ICAPS 2021 Workshop on
  Explainable AI Planning (XAIP)}}.
\newblock


\bibitem[Carroll et~al\mbox{.}(2019)]%
        {carroll2019utility}
\bibfield{author}{\bibinfo{person}{Micah Carroll}, \bibinfo{person}{Rohin
  Shah}, \bibinfo{person}{Mark~K Ho}, \bibinfo{person}{Tom Griffiths},
  \bibinfo{person}{Sanjit Seshia}, \bibinfo{person}{Pieter Abbeel}, {and}
  \bibinfo{person}{Anca Dragan}.} \bibinfo{year}{2019}\natexlab{}.
\newblock \showarticletitle{On the utility of learning about humans for
  human-ai coordination}.
\newblock \bibinfo{journal}{\emph{NeurIPS}} (\bibinfo{year}{2019}).
\newblock


\bibitem[Chakraborti et~al\mbox{.}(2020)]%
        {chakraborti2020emerging}
\bibfield{author}{\bibinfo{person}{Tathagata Chakraborti},
  \bibinfo{person}{Sarath Sreedharan}, {and} \bibinfo{person}{Subbarao
  Kambhampati}.} \bibinfo{year}{2020}\natexlab{}.
\newblock \showarticletitle{The Emerging Landscape of Explainable Automated
  Planning \& Decision Making.}. In \bibinfo{booktitle}{\emph{IJCAI}}.
  \bibinfo{pages}{4803--4811}.
\newblock


\bibitem[Chakraborti et~al\mbox{.}(2017)]%
        {chakraborti2017plan}
\bibfield{author}{\bibinfo{person}{Tathagata Chakraborti},
  \bibinfo{person}{Sarath Sreedharan}, \bibinfo{person}{Yu Zhang}, {and}
  \bibinfo{person}{Subbarao Kambhampati}.} \bibinfo{year}{2017}\natexlab{}.
\newblock \showarticletitle{Plan explanations as model reconciliation: Moving
  beyond explanation as soliloquy}.
\newblock \bibinfo{journal}{\emph{arXiv preprint arXiv:1701.08317}}
  (\bibinfo{year}{2017}).
\newblock


\bibitem[Csibra and Gergely(2007)]%
        {csibra2007obsessed}
\bibfield{author}{\bibinfo{person}{Gergely Csibra} {and}
  \bibinfo{person}{Gy{\"o}rgy Gergely}.} \bibinfo{year}{2007}\natexlab{}.
\newblock \showarticletitle{‘Obsessed with goals’: Functions and mechanisms
  of teleological interpretation of actions in humans}.
\newblock \bibinfo{journal}{\emph{Acta psychologica}} \bibinfo{volume}{124},
  \bibinfo{number}{1} (\bibinfo{year}{2007}), \bibinfo{pages}{60--78}.
\newblock


\bibitem[Czechowski et~al\mbox{.}(2021)]%
        {czechowski2021subgoal}
\bibfield{author}{\bibinfo{person}{Konrad Czechowski}, \bibinfo{person}{Tomasz
  Odrzyg{\'o}{\'z}d{\'z}}, \bibinfo{person}{Marek Zbysi{\'n}ski},
  \bibinfo{person}{Micha{\l} Zawalski}, \bibinfo{person}{Krzysztof Olejnik},
  \bibinfo{person}{Yuhuai Wu}, \bibinfo{person}{Lukasz Kucinski}, {and}
  \bibinfo{person}{Piotr Mi{\l}o{\'s}}.} \bibinfo{year}{2021}\natexlab{}.
\newblock \showarticletitle{Subgoal Search For Complex Reasoning Tasks}.
\newblock \bibinfo{journal}{\emph{NeurIPS}} (\bibinfo{year}{2021}).
\newblock


\bibitem[Das et~al\mbox{.}(2021)]%
        {das2021explainable}
\bibfield{author}{\bibinfo{person}{Devleena Das}, \bibinfo{person}{Siddhartha
  Banerjee}, {and} \bibinfo{person}{Sonia Chernova}.}
  \bibinfo{year}{2021}\natexlab{}.
\newblock \showarticletitle{Explainable ai for robot failures: Generating
  explanations that improve user assistance in fault recovery}. In
  \bibinfo{booktitle}{\emph{International Conference on Human-Robot
  Interaction}}. \bibinfo{pages}{351--360}.
\newblock


\bibitem[Das and Chernova(2020)]%
        {das2020leveraging}
\bibfield{author}{\bibinfo{person}{Devleena Das} {and} \bibinfo{person}{Sonia
  Chernova}.} \bibinfo{year}{2020}\natexlab{}.
\newblock \showarticletitle{Leveraging rationales to improve human task
  performance}. In \bibinfo{booktitle}{\emph{IUI}}. \bibinfo{pages}{510--518}.
\newblock


\bibitem[Doshi-Velez and Kim(2017)]%
        {doshi2017towards}
\bibfield{author}{\bibinfo{person}{Finale Doshi-Velez} {and}
  \bibinfo{person}{Been Kim}.} \bibinfo{year}{2017}\natexlab{}.
\newblock \showarticletitle{Towards a rigorous science of interpretable machine
  learning}.
\newblock \bibinfo{journal}{\emph{arXiv preprint arXiv:1702.08608}}
  (\bibinfo{year}{2017}).
\newblock


\bibitem[Ehsan et~al\mbox{.}(2019)]%
        {ehsan2019automated}
\bibfield{author}{\bibinfo{person}{Upol Ehsan}, \bibinfo{person}{Pradyumna
  Tambwekar}, \bibinfo{person}{Larry Chan}, \bibinfo{person}{Brent Harrison},
  {and} \bibinfo{person}{Mark~O Riedl}.} \bibinfo{year}{2019}\natexlab{}.
\newblock \showarticletitle{Automated rationale generation: a technique for
  explainable AI and its effects on human perceptions}. In
  \bibinfo{booktitle}{\emph{International Conference on Intelligent User
  Interfaces}}. \bibinfo{pages}{263--274}.
\newblock


\bibitem[Eyerich et~al\mbox{.}(2009)]%
        {eyerich2009using}
\bibfield{author}{\bibinfo{person}{Patrick Eyerich}, \bibinfo{person}{Robert
  Mattm{\"u}ller}, {and} \bibinfo{person}{Gabriele R{\"o}ger}.}
  \bibinfo{year}{2009}\natexlab{}.
\newblock \showarticletitle{Using the context-enhanced additive heuristic for
  temporal and numeric planning}. In \bibinfo{booktitle}{\emph{ICAPS}}.
\newblock


\bibitem[Feng et~al\mbox{.}(2020)]%
        {feng2020explainable}
\bibfield{author}{\bibinfo{person}{Jinyue Feng}, \bibinfo{person}{Chantal
  Shaib}, {and} \bibinfo{person}{Frank Rudzicz}.}
  \bibinfo{year}{2020}\natexlab{}.
\newblock \showarticletitle{Explainable clinical decision support from text}.
  In \bibinfo{booktitle}{\emph{EMNLP}}. \bibinfo{pages}{1478--1489}.
\newblock


\bibitem[Fox and Long(2003)]%
        {fox2003pddl2}
\bibfield{author}{\bibinfo{person}{Maria Fox} {and} \bibinfo{person}{Derek
  Long}.} \bibinfo{year}{2003}\natexlab{}.
\newblock \showarticletitle{PDDL2. 1: An extension to PDDL for expressing
  temporal planning domains}.
\newblock \bibinfo{journal}{\emph{Journal of artificial intelligence research}}
   \bibinfo{volume}{20} (\bibinfo{year}{2003}), \bibinfo{pages}{61--124}.
\newblock


\bibitem[Grover et~al\mbox{.}(2020)]%
        {grover2020radar}
\bibfield{author}{\bibinfo{person}{Sachin Grover}, \bibinfo{person}{Sailik
  Sengupta}, \bibinfo{person}{Tathagata Chakraborti},
  \bibinfo{person}{Aditya~Prasad Mishra}, {and} \bibinfo{person}{Subbarao
  Kambhampati}.} \bibinfo{year}{2020}\natexlab{}.
\newblock \showarticletitle{RADAR: automated task planning for proactive
  decision support}.
\newblock \bibinfo{journal}{\emph{Human--Computer Interaction}}
  \bibinfo{volume}{35}, \bibinfo{number}{5-6} (\bibinfo{year}{2020}),
  \bibinfo{pages}{387--412}.
\newblock


\bibitem[Guerlain et~al\mbox{.}(2000)]%
        {guerlain2000intelligent}
\bibfield{author}{\bibinfo{person}{Stephanie Guerlain},
  \bibinfo{person}{Donald~E Brown}, {and} \bibinfo{person}{Christina
  Mastrangelo}.} \bibinfo{year}{2000}\natexlab{}.
\newblock \showarticletitle{Intelligent decision support systems}. In
  \bibinfo{booktitle}{\emph{SMC}}, Vol.~\bibinfo{volume}{3}. IEEE.
\newblock


\bibitem[Guti{\'e}rrez et~al\mbox{.}(2019)]%
        {Gutierrez2019}
\bibfield{author}{\bibinfo{person}{Francisco Guti{\'e}rrez},
  \bibinfo{person}{Xavier Ochoa}, \bibinfo{person}{Karsten Seipp},
  \bibinfo{person}{Tom Broos}, {and} \bibinfo{person}{Katrien Verbert}.}
  \bibinfo{year}{2019}\natexlab{}.
\newblock \showarticletitle{Benefits and Trade-Offs of Different Model
  Representations in Decision Support Systems for Non-expert Users}. In
  \bibinfo{booktitle}{\emph{INTERACT}},
  \bibfield{editor}{\bibinfo{person}{David Lamas}, \bibinfo{person}{Fernando
  Loizides}, \bibinfo{person}{Lennart Nacke}, \bibinfo{person}{Helen Petrie},
  \bibinfo{person}{Marco Winckler}, {and} \bibinfo{person}{Panayiotis
  Zaphiris}} (Eds.). \bibinfo{publisher}{Springer International Publishing},
  \bibinfo{pages}{576--597}.
\newblock


\bibitem[Hoffmann and Magazzeni(2019)]%
        {hoffmann2019explainable}
\bibfield{author}{\bibinfo{person}{J{\"o}rg Hoffmann} {and}
  \bibinfo{person}{Daniele Magazzeni}.} \bibinfo{year}{2019}\natexlab{}.
\newblock \showarticletitle{Explainable AI planning (XAIP): overview and the
  case of contrastive explanation}.
\newblock \bibinfo{journal}{\emph{Reasoning Web: Explainable Artificial
  Intelligence}} (\bibinfo{year}{2019}), \bibinfo{pages}{277--282}.
\newblock


\bibitem[Iovino et~al\mbox{.}(2020)]%
        {iovino2020survey}
\bibfield{author}{\bibinfo{person}{Matteo Iovino}, \bibinfo{person}{Edvards
  Scukins}, \bibinfo{person}{Jonathan Styrud}, \bibinfo{person}{Petter
  {\"O}gren}, {and} \bibinfo{person}{Christian Smith}.}
  \bibinfo{year}{2020}\natexlab{}.
\newblock \showarticletitle{A survey of behavior trees in robotics and ai}.
\newblock \bibinfo{journal}{\emph{arXiv preprint arXiv:2005.05842}}
  (\bibinfo{year}{2020}).
\newblock


\bibitem[Jones et~al\mbox{.}(2019)]%
        {jones2019malfunction}
\bibfield{author}{\bibinfo{person}{Richard~W Jones}, \bibinfo{person}{James~E
  Mateer}, {and} \bibinfo{person}{Michael~J Harrison}.}
  \bibinfo{year}{2019}\natexlab{}.
\newblock \showarticletitle{Malfunction transparency in clinical decision
  support systems: a classification approach}. In
  \bibinfo{booktitle}{\emph{ICIEA}}. IEEE, \bibinfo{pages}{1354--1359}.
\newblock


\bibitem[Kaelbling and Lozano-P{\'e}rez(2010)]%
        {kaelbling2010hierarchical}
\bibfield{author}{\bibinfo{person}{Leslie~Pack Kaelbling} {and}
  \bibinfo{person}{Tom{\'a}s Lozano-P{\'e}rez}.}
  \bibinfo{year}{2010}\natexlab{}.
\newblock \showarticletitle{Hierarchical planning in the now}. In
  \bibinfo{booktitle}{\emph{Workshops at AAAI}}.
\newblock


\bibitem[Kim et~al\mbox{.}(2018)]%
        {kim2018interpretability}
\bibfield{author}{\bibinfo{person}{Been Kim}, \bibinfo{person}{Martin
  Wattenberg}, \bibinfo{person}{Justin Gilmer}, \bibinfo{person}{Carrie Cai},
  \bibinfo{person}{James Wexler}, \bibinfo{person}{Fernanda Viegas},
  {et~al\mbox{.}}} \bibinfo{year}{2018}\natexlab{}.
\newblock \showarticletitle{Interpretability beyond feature attribution:
  Quantitative testing with concept activation vectors (tcav)}. In
  \bibinfo{booktitle}{\emph{ICML}}. PMLR, \bibinfo{pages}{2668--2677}.
\newblock


\bibitem[Lin et~al\mbox{.}(2009)]%
        {lin2009decision}
\bibfield{author}{\bibinfo{person}{Chinho Lin}, \bibinfo{person}{Chun~Mei Lin},
  \bibinfo{person}{Binshan Lin}, {and} \bibinfo{person}{Ming-Chin Yang}.}
  \bibinfo{year}{2009}\natexlab{}.
\newblock \showarticletitle{A decision support system for improving doctors’
  prescribing behavior}.
\newblock \bibinfo{journal}{\emph{Expert Systems with Applications}}
  \bibinfo{volume}{36}, \bibinfo{number}{4} (\bibinfo{year}{2009}),
  \bibinfo{pages}{7975--7984}.
\newblock


\bibitem[Liu et~al\mbox{.}(2020)]%
        {liu2020planning}
\bibfield{author}{\bibinfo{person}{Yuxin Liu}, \bibinfo{person}{Qiguang Chen},
  \bibinfo{person}{Ke Jin}, {and} \bibinfo{person}{Zhanhao Zhang}.}
  \bibinfo{year}{2020}\natexlab{}.
\newblock \showarticletitle{Planning for Overcooked Game with PDDL}.
\newblock \bibinfo{journal}{\emph{International Core Journal of Engineering}}
  \bibinfo{volume}{6}, \bibinfo{number}{12} (\bibinfo{year}{2020}),
  \bibinfo{pages}{315--325}.
\newblock


\bibitem[Machado et~al\mbox{.}(2018)]%
        {machado2018use}
\bibfield{author}{\bibinfo{person}{Jessica~P Machado}, \bibinfo{person}{Xuan~T
  Lam}, {and} \bibinfo{person}{Jung-Wei Chen}.}
  \bibinfo{year}{2018}\natexlab{}.
\newblock \showarticletitle{Use of a clinical decision support tool for the
  management of traumatic dental injuries in the primary dentition by novice
  and expert clinicians}.
\newblock \bibinfo{journal}{\emph{Dental Traumatology}} \bibinfo{volume}{34},
  \bibinfo{number}{2} (\bibinfo{year}{2018}), \bibinfo{pages}{120--128}.
\newblock


\bibitem[Newell et~al\mbox{.}(1972)]%
        {newell1972human}
\bibfield{author}{\bibinfo{person}{Allen Newell},
  \bibinfo{person}{Herbert~Alexander Simon}, {et~al\mbox{.}}}
  \bibinfo{year}{1972}\natexlab{}.
\newblock \bibinfo{booktitle}{\emph{Human problem solving}}.
  Vol.~\bibinfo{volume}{104}.
\newblock \bibinfo{publisher}{Prentice-hall Englewood Cliffs, NJ}.
\newblock


\bibitem[Norvig and Intelligence(2002)]%
        {norvig2002modern}
\bibfield{author}{\bibinfo{person}{P~Russel Norvig} {and}
  \bibinfo{person}{S~Artificial Intelligence}.}
  \bibinfo{year}{2002}\natexlab{}.
\newblock \showarticletitle{A modern approach}.
\newblock \bibinfo{journal}{\emph{Prentice Hall Upper Saddle River, NJ, USA:
  Rani, M., Nayak, R., \& Vyas, OP (2015). An ontology-based adaptive
  personalized e-learning system, assisted by software agents on cloud storage.
  Knowledge-Based Systems}}  \bibinfo{volume}{90} (\bibinfo{year}{2002}),
  \bibinfo{pages}{33--48}.
\newblock


\bibitem[Nourani et~al\mbox{.}(2020)]%
        {nourani2020role}
\bibfield{author}{\bibinfo{person}{Mahsan Nourani}, \bibinfo{person}{Joanie
  King}, {and} \bibinfo{person}{Eric Ragan}.} \bibinfo{year}{2020}\natexlab{}.
\newblock \showarticletitle{The role of domain expertise in user trust and the
  impact of first impressions with intelligent systems}. In
  \bibinfo{booktitle}{\emph{AAAI Conference on Human Computation and
  Crowdsourcing}}, Vol.~\bibinfo{volume}{8}. \bibinfo{pages}{112--121}.
\newblock


\bibitem[Papamichail and French(2000)]%
        {papamichail2000decision}
\bibfield{author}{\bibinfo{person}{KN Papamichail} {and} \bibinfo{person}{S
  French}.} \bibinfo{year}{2000}\natexlab{}.
\newblock \showarticletitle{Decision support in nuclear emergencies}.
\newblock \bibinfo{journal}{\emph{J. of hazardous materials}}
  \bibinfo{volume}{71}, \bibinfo{number}{1-3} (\bibinfo{year}{2000}),
  \bibinfo{pages}{321--342}.
\newblock


\bibitem[Rathore et~al\mbox{.}(2018)]%
        {RATHORE2018920}
\bibfield{author}{\bibinfo{person}{Shailendra Rathore},
  \bibinfo{person}{Vincenzo Loia}, {and} \bibinfo{person}{Jong~Hyuk Park}.}
  \bibinfo{year}{2018}\natexlab{}.
\newblock \showarticletitle{SpamSpotter: An efficient spammer detection
  framework based on intelligent decision support system on Facebook}.
\newblock \bibinfo{journal}{\emph{Applied Soft Computing}}
  \bibinfo{volume}{67} (\bibinfo{year}{2018}), \bibinfo{pages}{920--932}.
\newblock


\bibitem[Ribeiro et~al\mbox{.}(2016)]%
        {ribeiro2016should}
\bibfield{author}{\bibinfo{person}{Marco~Tulio Ribeiro},
  \bibinfo{person}{Sameer Singh}, {and} \bibinfo{person}{Carlos Guestrin}.}
  \bibinfo{year}{2016}\natexlab{}.
\newblock \showarticletitle{" Why should I trust you?" Explaining the
  predictions of any classifier}. In \bibinfo{booktitle}{\emph{Proc. of the
  22nd ACM SIGKDD international conference on knowledge discovery and data
  mining}}. \bibinfo{pages}{1135--1144}.
\newblock


\bibitem[Richter et~al\mbox{.}(2008)]%
        {richter2008landmarks}
\bibfield{author}{\bibinfo{person}{Silvia Richter}, \bibinfo{person}{Malte
  Helmert}, {and} \bibinfo{person}{Matthias Westphal}.}
  \bibinfo{year}{2008}\natexlab{}.
\newblock \showarticletitle{Landmarks Revisited.}. In
  \bibinfo{booktitle}{\emph{AAAI}}, Vol.~\bibinfo{volume}{8}.
  \bibinfo{pages}{975--982}.
\newblock


\bibitem[Rodr{\'\i}guez et~al\mbox{.}(2020)]%
        {rodriguez2020intelligent}
\bibfield{author}{\bibinfo{person}{Germ{\'a}n~Gonz{\'a}lez Rodr{\'\i}guez},
  \bibinfo{person}{Jose~M Gonzalez-Cava}, {and} \bibinfo{person}{Juan
  Albino~M{\'e}ndez P{\'e}rez}.} \bibinfo{year}{2020}\natexlab{}.
\newblock \showarticletitle{An intelligent decision support system for
  production planning based on machine learning}.
\newblock \bibinfo{journal}{\emph{Journal of Intelligent Manufacturing}}
  \bibinfo{volume}{31}, \bibinfo{number}{5} (\bibinfo{year}{2020}),
  \bibinfo{pages}{1257--1273}.
\newblock


\bibitem[Schank and Abelson(2013)]%
        {schank2013scripts}
\bibfield{author}{\bibinfo{person}{Roger~C Schank} {and}
  \bibinfo{person}{Robert~P Abelson}.} \bibinfo{year}{2013}\natexlab{}.
\newblock \bibinfo{booktitle}{\emph{Scripts, plans, goals, and understanding:
  An inquiry into human knowledge structures}}.
\newblock \bibinfo{publisher}{Psychology Press}.
\newblock


\bibitem[Seegebarth et~al\mbox{.}(2012)]%
        {seegebarth2012making}
\bibfield{author}{\bibinfo{person}{Bastian Seegebarth}, \bibinfo{person}{Felix
  M{\"u}ller}, \bibinfo{person}{Bernd Schattenberg}, {and}
  \bibinfo{person}{Susanne Biundo}.} \bibinfo{year}{2012}\natexlab{}.
\newblock \showarticletitle{Making hybrid plans more clear to human users-a
  formal approach for generating sound explanations}. In
  \bibinfo{booktitle}{\emph{Twenty-second international conference on automated
  planning and scheduling}}.
\newblock


\bibitem[Sreedharan et~al\mbox{.}(2019)]%
        {sreedharan2019can}
\bibfield{author}{\bibinfo{person}{Sarath Sreedharan},
  \bibinfo{person}{Siddharth Srivastava}, \bibinfo{person}{David Smith}, {and}
  \bibinfo{person}{Subbarao Kambhampati}.} \bibinfo{year}{2019}\natexlab{}.
\newblock \showarticletitle{Why Can't You Do That HAL? Explaining Unsolvability
  of Planning Tasks}. In \bibinfo{booktitle}{\emph{IJCAI}}.
\newblock


\bibitem[Sutton et~al\mbox{.}(2020)]%
        {sutton2020overview}
\bibfield{author}{\bibinfo{person}{Reed~T Sutton}, \bibinfo{person}{David
  Pincock}, \bibinfo{person}{Daniel~C Baumgart}, \bibinfo{person}{Daniel~C
  Sadowski}, \bibinfo{person}{Richard~N Fedorak}, {and}
  \bibinfo{person}{Karen~I Kroeker}.} \bibinfo{year}{2020}\natexlab{}.
\newblock \showarticletitle{An overview of clinical decision support systems:
  benefits, risks, and strategies for success}.
\newblock \bibinfo{journal}{\emph{NPJ digital medicine}} \bibinfo{volume}{3},
  \bibinfo{number}{1} (\bibinfo{year}{2020}), \bibinfo{pages}{1--10}.
\newblock


\bibitem[Tabrez et~al\mbox{.}(2019)]%
        {tabrez2019explanation}
\bibfield{author}{\bibinfo{person}{Aaquib Tabrez}, \bibinfo{person}{Shivendra
  Agrawal}, {and} \bibinfo{person}{Bradley Hayes}.}
  \bibinfo{year}{2019}\natexlab{}.
\newblock \showarticletitle{Explanation-based reward coaching to improve human
  performance via reinforcement learning}. In \bibinfo{booktitle}{\emph{HRI}}.
  IEEE, \bibinfo{pages}{249--257}.
\newblock


\bibitem[Vallacher and Wegner(1987)]%
        {vallacher1987people}
\bibfield{author}{\bibinfo{person}{Robin~R Vallacher} {and}
  \bibinfo{person}{Daniel~M Wegner}.} \bibinfo{year}{1987}\natexlab{}.
\newblock \showarticletitle{What do people think they're doing? Action
  identification and human behavior.}
\newblock \bibinfo{journal}{\emph{Psychological review}} \bibinfo{volume}{94},
  \bibinfo{number}{1} (\bibinfo{year}{1987}), \bibinfo{pages}{3}.
\newblock


\bibitem[Valmeekam et~al\mbox{.}(2020)]%
        {valmeekam2020radar}
\bibfield{author}{\bibinfo{person}{Karthik Valmeekam}, \bibinfo{person}{Sarath
  Sreedharan}, \bibinfo{person}{Sailik Sengupta}, {and}
  \bibinfo{person}{Subbarao Kambhampati}.} \bibinfo{year}{2020}\natexlab{}.
\newblock \showarticletitle{RADAR-X: An Interactive Interface Pairing
  Contrastive Explanations with Revised Plan Suggestions}.
\newblock \bibinfo{journal}{\emph{arXiv preprint arXiv:2011.09644}}
  (\bibinfo{year}{2020}).
\newblock


\bibitem[Walsh et~al\mbox{.}(2019)]%
        {walsh2019decision}
\bibfield{author}{\bibinfo{person}{Se{\'a}n Walsh}, \bibinfo{person}{Evelyn~EC
  de Jong}, \bibinfo{person}{Janna~E van Timmeren}, \bibinfo{person}{Abdalla
  Ibrahim}, \bibinfo{person}{Inge Compter}, \bibinfo{person}{Jurgen Peerlings},
  \bibinfo{person}{Sebastian Sanduleanu}, \bibinfo{person}{Turkey Refaee},
  \bibinfo{person}{Simon Keek}, \bibinfo{person}{Ruben~THM Larue},
  {et~al\mbox{.}}} \bibinfo{year}{2019}\natexlab{}.
\newblock \showarticletitle{Decision support systems in oncology}.
\newblock \bibinfo{journal}{\emph{JCO clinical cancer informatics}}
  \bibinfo{volume}{3} (\bibinfo{year}{2019}), \bibinfo{pages}{1--9}.
\newblock


\bibitem[Wu et~al\mbox{.}(2021)]%
        {wu2021too}
\bibfield{author}{\bibinfo{person}{Sarah~A Wu}, \bibinfo{person}{Rose~E Wang},
  \bibinfo{person}{James~A Evans}, \bibinfo{person}{Joshua~B Tenenbaum},
  \bibinfo{person}{David~C Parkes}, {and} \bibinfo{person}{Max
  Kleiman-Weiner}.} \bibinfo{year}{2021}\natexlab{}.
\newblock \showarticletitle{Too Many Cooks: Bayesian Inference for Coordinating
  Multi-Agent Collaboration}.
\newblock \bibinfo{journal}{\emph{Topics in Cognitive Science}}
  \bibinfo{volume}{13}, \bibinfo{number}{2} (\bibinfo{year}{2021}),
  \bibinfo{pages}{414--432}.
\newblock


\bibitem[Zhang et~al\mbox{.}(2019)]%
        {zhang2019interpreting}
\bibfield{author}{\bibinfo{person}{Quanshi Zhang}, \bibinfo{person}{Yu Yang},
  \bibinfo{person}{Haotian Ma}, {and} \bibinfo{person}{Ying~Nian Wu}.}
  \bibinfo{year}{2019}\natexlab{}.
\newblock \showarticletitle{Interpreting cnns via decision trees}. In
  \bibinfo{booktitle}{\emph{Proceedings of the IEEE Conference on Computer
  Vision and Pattern Recognition}}. \bibinfo{pages}{6261--6270}.
\newblock


\bibitem[Zhang et~al\mbox{.}(2017)]%
        {zhang2017plan}
\bibfield{author}{\bibinfo{person}{Yu Zhang}, \bibinfo{person}{Sarath
  Sreedharan}, \bibinfo{person}{Anagha Kulkarni}, \bibinfo{person}{Tathagata
  Chakraborti}, \bibinfo{person}{Hankz~Hankui Zhuo}, {and}
  \bibinfo{person}{Subbarao Kambhampati}.} \bibinfo{year}{2017}\natexlab{}.
\newblock \showarticletitle{Plan explicability and predictability for robot
  task planning}. In \bibinfo{booktitle}{\emph{2017 IEEE international
  conference on robotics and automation (ICRA)}}. IEEE,
  \bibinfo{pages}{1313--1320}.
\newblock


\bibitem[Zhuang et~al\mbox{.}(2009)]%
        {zhuang2009combining}
\bibfield{author}{\bibinfo{person}{Zoe~Y Zhuang}, \bibinfo{person}{Leonid
  Churilov}, \bibinfo{person}{Frada Burstein}, {and} \bibinfo{person}{Ken
  Sikaris}.} \bibinfo{year}{2009}\natexlab{}.
\newblock \showarticletitle{Combining data mining and case-based reasoning for
  intelligent decision support for pathology ordering by general
  practitioners}.
\newblock \bibinfo{journal}{\emph{European Journal of Operational Research}}
  \bibinfo{volume}{195}, \bibinfo{number}{3} (\bibinfo{year}{2009}),
  \bibinfo{pages}{662--675}.
\newblock


\end{thebibliography}
\end{document}